\documentclass[11pt]{article}

\usepackage[margin=1in]{geometry}
\usepackage{times}
\usepackage[T1]{fontenc}
\usepackage[utf8]{inputenc}
\usepackage{microtype}
\usepackage{titlesec}
\usepackage{float}
\usepackage{graphicx}
\usepackage{subcaption}
\usepackage{caption}
\usepackage{multirow}
\usepackage{array}
\usepackage{amsmath}
\usepackage{amssymb}
 \usepackage{tikz}
 \usetikzlibrary{shapes.geometric, arrows.meta, positioning, fit, backgrounds}
\usepackage{url}
\usepackage[main=english]{babel}
\usepackage{tikz}
\usetikzlibrary{shapes.geometric, arrows.meta, positioning}

\usepackage[
  colorlinks=true,
  linkcolor=blue,
  citecolor=blue,
  urlcolor=blue,
  filecolor=blue,
  pdftitle={Benchmarking Linguistic Adaptation in Comparable-Sized LLMs},
  pdfauthor={Ananda Rimal, Adarsha Rimal}
]{hyperref}
\usepackage{cite}   % compressed [1,2,3] style; load AFTER hyperref

\titleformat{\section}{\large\bfseries\MakeUppercase}{\thesection}{1em}{}
\titleformat{\subsection}{\normalsize\bfseries}{\thesubsection}{1em}{}
\titleformat{\subsubsection}{\normalsize\itshape}{\thesubsubsection}{1em}{}

\title{
    \vspace{-0.8in}
    \hrule height 1.5pt
    \vspace{0.2in}
    \textbf{\Large BENCHMARKING LINGUISTIC ADAPTATION IN
    COMPARABLE-SIZED LLMs: A STUDY OF LLAMA-3.1-8B,
    MISTRAL-7B-v0.1, AND QWEN3-8B ON ROMANIZED NEPALI}
    \vspace{0.2in}
    \hrule height 1.5pt
    \vspace{0.2in}
}

\author{
    \begin{minipage}[t]{0.45\textwidth}
        \centering
        \textbf{Ananda Rimal} \\
        \small Dept.\ of Computer Science \& Engineering \\
        Nepal Engineering College \\
        \texttt{anandr022342@nec.edu.np}
    \end{minipage}
    \begin{minipage}[t]{0.45\textwidth}
        \centering
        \textbf{Adarsha Rimal} \\
        \small Central Dept.\ of CS and IT \\
        Tribhuvan University \\
        \texttt{adarsharimal07@gmail.com}
    \end{minipage}
}

\date{}
\usepackage{longtable}

\begin{document}
\maketitle

% ── Abstract 
% ── Abstract ────────────────────────────────────────────────
% ── Abstract ────────────────────────────────────────────────
\begin{center}
    \textbf{\normalsize\MakeUppercase{Abstract}}
\end{center}
\begin{center}
\begin{minipage}{0.9\textwidth}
\noindent
Romanized Nepali, the Nepali language written in the
Latin alphabet, is the dominant medium for informal
digital communication in Nepal, yet it remains critically
underresourced in the landscape of Large Language Models
(LLMs). This study presents a systematic benchmarking of
linguistic adaptation across three comparable-sized
open-weight models: Llama-3.1-8B, Mistral-7B-v0.1, and
Qwen3-8B. We evaluate these architectures under zero-shot
and fine-tuned settings using a curated bilingual dataset
of 10,000 transliterated instruction-following samples.
Performance is quantified across five metrics spanning
seven measurement dimensions: Perplexity (PPL),
BERTScore, chrF++, ROUGE-1, ROUGE-2, ROUGE-L, and BLEU,
capturing fluency, phonetic consistency, and semantic
integrity.
Models were fine-tuned using Quantized Low-Rank Adaptation
(QLoRA)~\cite{dettmers2024qlora} with Rank-Stabilized
LoRA (rsLoRA)~\cite{kalajdzievski2023rslora} at rank
$r=32$ on dual NVIDIA Tesla T4 GPUs, training only
$\approx 1\%$ of each model's parameters in under 27
total GPU-hours. At zero-shot, all three models fail to
generate Romanized Nepali, each exhibiting a distinct
architecture-specific failure mode. Following fine-tuning,
all three resolve these failures and converge to
BERTScore $\approx 0.75$ and chrF++ $> 23$. Overall
dimension-wise assessment across ten criteria identifies
Qwen3-8B as the overall recommended architecture, being
the only model to produce semantically relevant zero-shot
output and leading all structural alignment metrics
post-SFT. The \textit{adaptation headroom hypothesis} is
confirmed: Llama-3.1-8B, despite its weakest zero-shot
baseline, achieves the largest absolute fine-tuning gains
in PPL ($\Delta = -49.77$) and BERTScore
($\Delta = +0.3287$), making it the preferred choice for
iterative low-resource development pipelines. This work
establishes the first rigorous baseline for Romanized
Nepali adaptation in comparable-sized open-weight LLMs.
\end{minipage}
\end{center}

\vspace{0.1in}

% ════════════════════════════════════════════════════════════
\section{Introduction}
% ════════════════════════════════════════════════════════════

The proliferation of social media and mobile messaging platforms
has established Romanized Nepali the Nepali language written
using the Latin alphabet as the de facto standard for
informal digital discourse in Nepal. Despite its ubiquity,
Romanized Nepali remains significantly under-resourced in
Natural Language Processing. Unlike formal Devanagari script,
Romanized Nepali lacks a standardized orthography, leading to
extreme phonetic variation where a single word may be
transliterated in multiple ways, such as ``khana,'' and 
``khaana,''~\cite{shahi2022review}.

Nepali is the official language of Nepal, spoken by approximately
44.86\% of the population~\cite{census2021nepal}. The language
is traditionally written in Devanagari script, comprising 36
consonants and 13 vowels. Current LLMs are predominantly trained
on standardized scripts and formal datasets, leaving transliterated
variants in a ``linguistic shadow''~\cite{bender2021dangers}.
This gap results in poor performance on downstream tasks such as
sentiment analysis, machine translation, and conversational AI
in the Nepali digital context.

This paper investigates the robustness and adaptation capabilities
of the comparable-sized LLMs, specifically Llama-3.1-8B~\cite{touvron2023llama},
Mistral-7B-v0.1~\cite{jiang2023mistral}, and
Qwen3-8B~\cite{bai2023qwen}. We focus on this size class due
to its increasing importance for on device deployment and
resource efficient fine tuning in low resource environments.
Through a systematic benchmark, we evaluate how different
architectures handle the phonetic noise and non standard syntax
inherent in Romanized Nepali, providing a foundation for future
generative AI applications for the Nepali diaspora.

To the best of our knowledge, this is the first comprehensive
study focusing specifically on the transliteration resilience
of comparable-sized LLMs on Romanized Nepali script.

% ════════════════════════════════════════════════════════════
\section{Literature Review}
% ════════════════════════════════════════════════════════════

\subsection{Transformer-Based Language Models}

The transformer architecture introduced by Vaswani et al.\
\cite{vaswani2017attention} marked a fundamental shift in NLP
by replacing recurrence with self-attention, enabling efficient
parallel processing and long-range dependency modeling. This
underpins virtually all modern LLMs. Brown et al.\
\cite{brown2020gpt3} demonstrated that scaling transformer
models to billions of parameters yields emergent few-shot and
zero-shot generalization, establishing the paradigm of
large-scale pre-training followed by lightweight task-specific
adaptation that all three model families in this study follow.

\subsection{Open-Weight Models in the 7--8B Class}

Recent open-weight LLMs have made high-performance language
modeling accessible for low-resource research. Touvron et al.\
\cite{touvron2023llama} introduced the LLaMA family,
demonstrating competitive performance at 7--65B parameters
through careful data curation. Jiang et al.\
\cite{jiang2023mistral} released Mistral 7B, incorporating
grouped-query and sliding window attention for improved
inference efficiency. Bai et al.\ \cite{bai2023qwen} presented
the Qwen series with broader multilingual pre-training coverage,
providing stronger cross-lingual transfer to non-English
scripts. These three model families form the subject of the
present study due to their comparable size ($\approx$7--8B
parameters), open-weight availability, and suitability for
on-device deployment.

\subsection{Nepali NLP: History and Current State}

Natural Language Processing for Nepali has a relatively short
but accelerating history. Early work focused on rule based and
statistical approaches to morphological analysis and
part-of-speech tagging, motivated by Nepali's highly
agglutinative morphology and complex sandhi rules that
complicate standard tokenization~\cite{shahi2022review}.
Bam and Shahi~\cite{bam2014nepali} demonstrated that even
basic preprocessing such as stopword removal and stemming
requires language specific adaptation, as standard multilingual
tools fail to account for Devanagari's orthographic properties.

Koirala and Niraula~\cite{koirala2021npvec}
introduced word embedding models for Nepali, establishing
distributional semantic resources for formal text but
explicitly noting that Romanized Nepali remains outside their
coverage.

Sitaula et al.\ \cite{sitaula2021napit} adapted multilingual
BERT via continued pre-training on a Devanagari corpus to
produce NepBERT, showing strong performance on formal NLP
tasks. Most recently, Pudasaini et al.\
\cite{pudasaini2025nepaligpt} developed NepaliGPT, the first
autoregressive generative model for Nepali, again focused
exclusively on Devanagari. Across this body of work, informal
Romanized Nepali has remained consistently unaddressed 
the principal gap this study targets.

\subsection{Tokenization for Low-Resource and Romanized Scripts}

Tokenization is a foundational bottleneck for LLM performance
on low-resource languages. Rust et al.\ \cite{rust2021good}
conducted a systematic evaluation of multilingual BERT
tokenizers across 99 languages, demonstrating that tokenizer
quality is a strong predictor of downstream task performance.
Languages with high subword fertility experience longer
effective sequence lengths, reduced context window utility,
and higher inference cost~\cite{rust2021good}.

Mielke et al.\ \cite{mielke2021between} showed that
agglutinative languages and those with non-Latin scripts are
disproportionately harmed by BPE-based vocabularies trained
on predominantly English corpora. Romanized Nepali faces a
compounded version of this problem: it is neither standard
English nor Devanagari, placing it in a tokenization dead zone
between the two primary training distributions of all three
studied models.

Kudo and Richardson~\cite{kudo2018sentencepiece} introduced
SentencePiece, a language-agnostic subword tokenizer that
treats text as a raw character sequence without whitespace
assumptions, giving it better generalization to unseen scripts.
Mistral-7B-v0.1 and Qwen3-8B both use
SentencePiece~\cite{jiang2023mistral, bai2023qwen}, while
Llama-3.1-8B uses Tiktoken~\cite{touvron2023llama}, a BPE
implementation optimized for English-heavy data. The zero-shot
failure analysis in this study directly reflects these
tokenizer design differences.

\subsection{Parameter-Efficient Fine-Tuning}

Full fine-tuning of LLMs is computationally prohibitive in
low-resource settings. Hu et al.\ \cite{hu2022lora} introduced
LoRA, which injects trainable low-rank matrices into frozen
transformer weights, reducing trainable parameters to under 1\%
of the base model. Dettmers et al.\ \cite{dettmers2024qlora}
extended this with QLoRA, combining 4-bit NF4 quantization
with LoRA adapter training, enabling fine-tuning on
consumer-grade hardware. Kalajdzievski~\cite{kalajdzievski2023rslora}
identified rank-dependent scaling instability in standard LoRA
and proposed rsLoRA, replacing the $\alpha/r$ scaling factor
with $\alpha/\sqrt{r}$ to stabilize training at $r=32$, the
rank used in this work.

Instruction fine-tuning~\cite{wei2021finetuned} has been shown
to be particularly effective for generalization to new task
formats, training models to follow structured
Instruction-Input-Output templates. The Alpaca dataset
format~\cite{taori2023alpaca} used in this study is a widely
adopted instantiation of this approach.

\subsection{Evaluation Metrics for Low-Resource Generation}

Papineni et al.\ \cite{papineni2002bleu} introduced BLEU,
the dominant $n$-gram precision metric for machine
translation, though its reliance on exact surface-form
matches makes it poorly suited to non-standardized
scripts~\cite{post2018call}. Popovic~\cite{popovic2015chrf}
proposed chrF, a character $n$-gram F-score more robust to
spelling variation. Zhang et al.\ \cite{zhang2019bertscore}
introduced BERTScore, evaluating semantic similarity via
contextual embeddings. Perplexity, grounded in information
theory~\cite{jurafsky2023speech}, measures predictive
confidence and serves as the primary checkpoint selection
criterion in this work. Lin~\cite{lin2004rouge} proposed
ROUGE-L for structural alignment via longest common
subsequence matching.

\subsection{Research Gap}

Across the reviewed literature three gaps are evident. First,
all existing Nepali NLP work targets the formal Devanagari
script, leaving informal Romanized Nepali
unaddressed~\cite{shahi2022review, pudasaini2025nepaligpt,
sitaula2021napit}. Second, while tokenizer quality is known
to predict downstream performance for low-resource
scripts~\cite{rust2021good, mielke2021between}, no study
has compared tokenizer design across competing comparable-sized
LLM architectures for a non-standardized transliterated script.
Third, no prior work has benchmarked the adaptation of
comparable-sized open-weight LLMs to Romanized Nepali under a
rigorous multi-metric framework. This study addresses all
three gaps.

% ════════════════════════════════════════════════════════════
% ════════════════════════════════════════════════════════════
\section{Methodology}
% ════════════════════════════════════════════════════════════

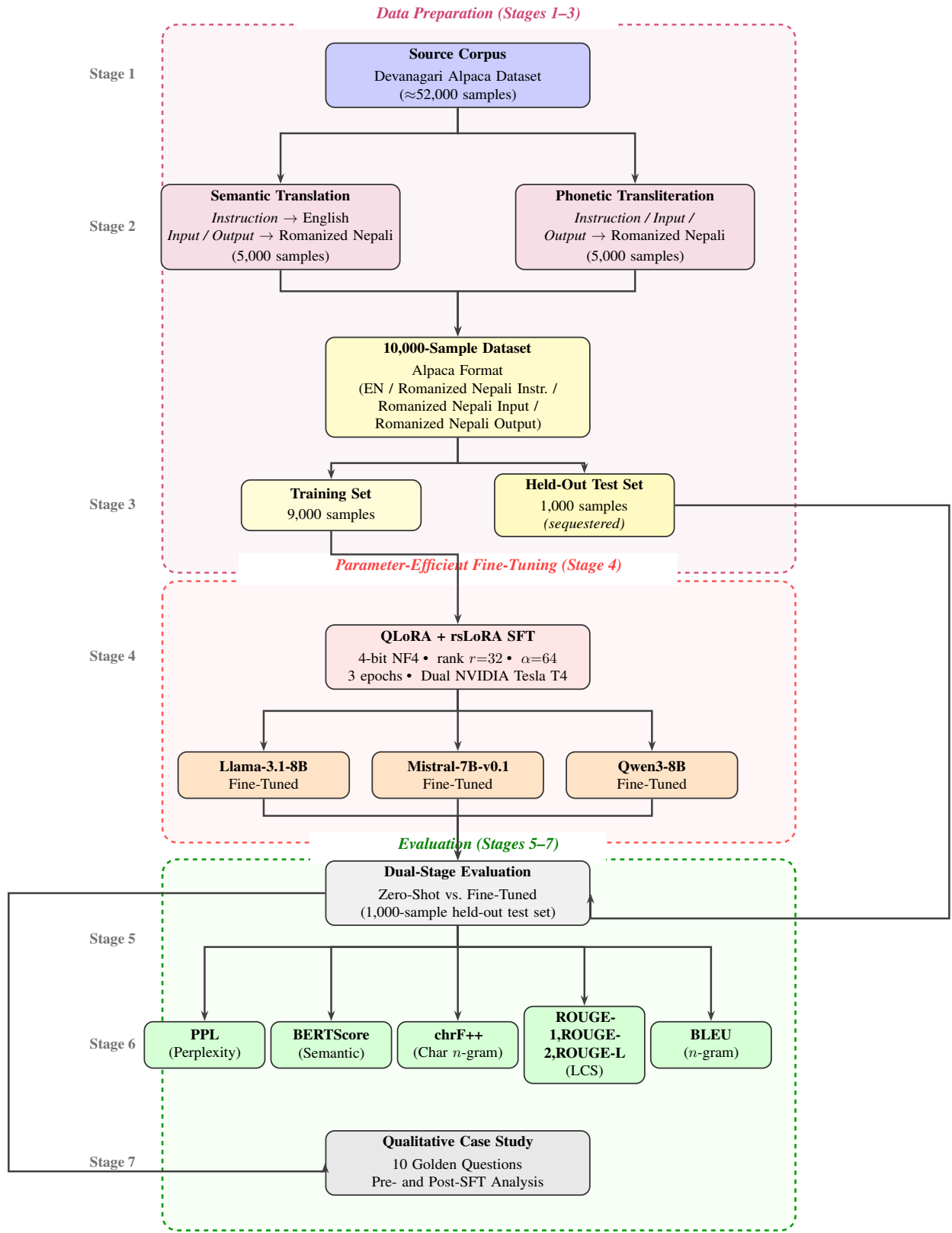
\begin{figure}[H]
\centering
\resizebox{\textwidth}{!}{%
\begin{tikzpicture}[
    mainbox/.style={
        rectangle, rounded corners=6pt,
        draw=black!80, line width=0.9pt,
        fill=blue!15,
        text width=6.0cm,
        align=center,
        minimum height=1.3cm,
        font=\small
    },
    databox/.style={
        rectangle, rounded corners=6pt,
        draw=black!80, line width=0.9pt,
        fill=purple!13,
        text width=5.4cm,
        align=center,
        minimum height=1.6cm,
        font=\small
    },
    mergedbox/.style={
        rectangle, rounded corners=6pt,
        draw=black!80, line width=0.9pt,
        fill=yellow!25,
        text width=6.0cm,
        align=center,
        minimum height=1.3cm,
        font=\small
    },
    splitbox/.style={
        rectangle, rounded corners=6pt,
        draw=black!80, line width=0.9pt,
        fill=yellow!18,
        text width=4.0cm,
        align=center,
        minimum height=1.2cm,
        font=\small
    },
    finetuneBox/.style={
        rectangle, rounded corners=6pt,
        draw=black!80, line width=0.9pt,
        fill=red!10,
        text width=6.0cm,
        align=center,
        minimum height=1.3cm,
        font=\small
    },
    modelbox/.style={
        rectangle, rounded corners=6pt,
        draw=black!80, line width=0.9pt,
        fill=orange!22,
        text width=3.8cm,
        align=center,
        minimum height=1.1cm,
        font=\small
    },
    evalbox/.style={
        rectangle, rounded corners=6pt,
        draw=black!80, line width=0.9pt,
        fill=gray!12,
        text width=6.0cm,
        align=center,
        minimum height=1.3cm,
        font=\small
    },
    metricbox/.style={
        rectangle, rounded corners=6pt,
        draw=black!80, line width=0.9pt,
        fill=green!16,
        text width=2.6cm,
        align=center,
        minimum height=1.1cm,
        font=\small
    },
    qualbox/.style={
        rectangle, rounded corners=6pt,
        draw=black!80, line width=0.9pt,
        fill=gray!18,
        text width=6.0cm,
        align=center,
        minimum height=1.3cm,
        font=\small
    },
    arrow/.style={
        -{Stealth[length=7pt, width=5pt]},
        line width=1.3pt,
        black!75
    },
    stagelabel/.style={
        font=\small\bfseries,
        text=black!55,
        anchor=east
    }
]

% ════════════════════════════════════════════════════
% BACKGROUND RECTANGLES
% ════════════════════════════════════════════════════
\begin{pgfonlayer}{background}

    % Data Preparation (Stages 1–3)
    \fill[purple!4, rounded corners=10pt]
        (-7.0, 1.2) rectangle (8.0, -11.8);
    \draw[purple!65, dashed, line width=1.3pt, rounded corners=10pt]
        (-7.0, 1.2) rectangle (8.0, -11.8);
    \fill[white] (-3.8, 1.1) rectangle (3.8, 1.75);
    \node[font=\normalsize\itshape\bfseries, text=purple!75]
        at (0.5, 1.45) {\textit{Data Preparation (Stages 1--3)}};

    % Parameter-Efficient Fine-Tuning (Stage 4)
    \fill[red!3, rounded corners=10pt]
        (-7.0, -12.0) rectangle (8.0, -18.2);
    \draw[red!65, dashed, line width=1.3pt, rounded corners=10pt]
        (-7.0, -12.0) rectangle (8.0, -18.2);
    \fill[white] (-5.2, -12.1) rectangle (5.2, -11.35);
    \node[font=\normalsize\itshape\bfseries, text=red!75]
        at (0.5, -11.65)
        {\textit{Parameter-Efficient Fine-Tuning (Stage 4)}};

    % Evaluation (Stages 5–7)
    \fill[green!3, rounded corners=10pt]
        (-7.0, -18.6) rectangle (8.0, -27.4);
    \draw[green!60!black, dashed, line width=1.3pt, rounded corners=10pt]
        (-7.0, -18.6) rectangle (8.0, -27.4);
    \fill[white] (-3.5, -18.7) rectangle (3.5, -17.95);
    \node[font=\normalsize\itshape\bfseries, text=green!50!black]
        at (0.5, -18.25)
        {\textit{Evaluation (Stages 5--7)}};

\end{pgfonlayer}

% ════════════════════════════════════════════════════
% STAGE LABELS — x=-7.5, always outside box boundary
% ════════════════════════════════════════════════════
\node[stagelabel] at (-7.5,  0.0)  {Stage 1};
\node[stagelabel] at (-7.5, -3.6)  {Stage 2};
\node[stagelabel] at (-7.5, -10.2) {Stage 3};
\node[stagelabel] at (-7.5, -13.8) {Stage 4};
\node[stagelabel] at (-7.5, -20.5) {Stage 5};
\node[stagelabel] at (-7.5, -23.0) {Stage 6};
\node[stagelabel] at (-7.5, -25.8) {Stage 7};

% ════════════════════════════════════════════════════
% STAGE 1 — Source Corpus
% ════════════════════════════════════════════════════
\node[mainbox, fill=blue!20] (source) at (0, 0)
    {\textbf{Source Corpus}\\[3pt]
     Devanagari Alpaca Dataset\\
     ($\approx$52{,}000 samples)};

% ════════════════════════════════════════════════════
% STAGE 2 — Two parallel branches
% ════════════════════════════════════════════════════
\node[databox] (translate) at (-4.2, -3.6)
    {\textbf{Semantic Translation}\\[3pt]
     \textit{Instruction} $\rightarrow$ English\\
     \textit{Input\;/\;Output} $\rightarrow$ Romanized Nepali\\[2pt]
     (5{,}000 samples)};

\node[databox] (translit) at (+4.2, -3.6)
    {\textbf{Phonetic Transliteration}\\[3pt]
     \textit{Instruction\;/\;Input\;/}\\
     \textit{Output} $\rightarrow$ Romanized Nepali\\[2pt]
     (5{,}000 samples)};

\draw[arrow] (source.south) -- ++(0,-0.6) -| (translate.north);
\draw[arrow] (source.south) -- ++(0,-0.6) -| (translit.north);

% ════════════════════════════════════════════════════
% MERGED DATASET
% ════════════════════════════════════════════════════
\node[mergedbox] (dataset) at (0, -7.4)
    {\textbf{10{,}000-Sample Dataset}\\[3pt]
     Alpaca Format\\
     (EN / Romanized Nepali Instr.\ /\\
      Romanized Nepali Input /\\
      Romanized Nepali Output)};

\draw[arrow] (translate.south) -- ++(0,-0.5) -| (dataset.north);
\draw[arrow] (translit.south)  -- ++(0,-0.5) -| (dataset.north);

% ════════════════════════════════════════════════════
% STAGE 3 — Train / Test Split
% ════════════════════════════════════════════════════
\node[splitbox] (train) at (-3.0, -10.2)
    {\textbf{Training Set}\\[2pt]
     9{,}000 samples};

\node[splitbox, fill=yellow!30] (test) at (+3.0, -10.2)
    {\textbf{Held-Out Test Set}\\[2pt]
     1{,}000 samples\\
     \small\textit{(sequestered)}};

\draw[arrow] (dataset.south) -- ++(0,-0.6) -| (train.north);
\draw[arrow] (dataset.south) -- ++(0,-0.6) -| (test.north);

% ════════════════════════════════════════════════════
% STAGE 4 — QLoRA + rsLoRA SFT
% ════════════════════════════════════════════════════
\node[finetuneBox] (qlora) at (0, -13.8)
    {\textbf{QLoRA + rsLoRA SFT}\\[3pt]
     4-bit NF4\;\textbullet\;
     rank $r{=}32$\;\textbullet\;
     $\alpha{=}64$\\
     3 epochs\;\textbullet\;
     Dual NVIDIA Tesla T4};

\draw[arrow] (train.south) -- ++(0,-0.55) -| (qlora.north);

\node[modelbox] (llama) at (-4.6, -16.6)
    {\textbf{Llama-3.1-8B}\\Fine-Tuned};
\node[modelbox] (mistral) at (0, -16.6)
    {\textbf{Mistral-7B-v0.1}\\Fine-Tuned};
\node[modelbox] (qwen) at (+4.6, -16.6)
    {\textbf{Qwen3-8B}\\Fine-Tuned};

\draw[arrow] (qlora.south) -- ++(0,-0.5) -| (llama.north);
\draw[arrow] (qlora.south) -- ++(0,-0.5)  -- (mistral.north);
\draw[arrow] (qlora.south) -- ++(0,-0.5) -| (qwen.north);

% ════════════════════════════════════════════════════
% STAGE 5 — Dual-Stage Evaluation (moved up to -19.4)
% ════════════════════════════════════════════════════
\node[evalbox] (eval) at (0, -19.4)
    {\textbf{Dual-Stage Evaluation}\\[3pt]
     Zero-Shot vs.\ Fine-Tuned\\
     (1{,}000-sample held-out test set)};

% Arrows from 3 models into eval with visible arrowheads
\draw[arrow] (llama.south)   -- ++(0,-0.4) -| (eval.north);
\draw[arrow] (mistral.south) -- ++(0,-0.4)  -- (eval.north);
\draw[arrow] (qwen.south)    -- ++(0,-0.4) -| (eval.north);

% ════════════════════════════════════════════════════
% SOLID arrow: Held-Out Test Set → Dual-Stage Eval
% Exits test.east, goes right then down, enters eval.east
% ════════════════════════════════════════════════════
\draw[arrow]
    (test.east)
    -- ++(6.5, 0)
    -- ++(0, -9.8)
    -| (eval.east);

% ════════════════════════════════════════════════════
% STAGE 6 — Five Metrics
% ════════════════════════════════════════════════════
\node[metricbox] (ppl)   at (-6.0, -23.0)
    {\textbf{PPL}\\(Perplexity)};
\node[metricbox] (bert)  at (-3.0, -23.0)
    {\textbf{BERTScore}\\(Semantic)};
\node[metricbox] (chrf)  at ( 0.0, -23.0)
    {\textbf{chrF++}\\(Char $n$-gram)};
\node[metricbox] (rouge) at (+3.0, -23.0)
    {\textbf{ROUGE-1,ROUGE-2,ROUGE-L}\\(LCS)};
\node[metricbox] (bleu)  at (+6.0, -23.0)
    {\textbf{BLEU}\\($n$-gram)};

\draw[arrow] (eval.south) -- ++(0,-0.5) -| (ppl.north);
\draw[arrow] (eval.south) -- ++(0,-0.5) -| (bert.north);
\draw[arrow] (eval.south) -- ++(0,-0.65) -- (chrf.north);
\draw[arrow] (eval.south) -- ++(0,-0.5) -| (rouge.north);
\draw[arrow] (eval.south) -- ++(0,-0.5) -| (bleu.north);

% ════════════════════════════════════════════════════
% STAGE 7 — Qualitative Case Study
% ════════════════════════════════════════════════════
\node[qualbox] (qual) at (0, -25.8)
    {\textbf{Qualitative Case Study}\\[3pt]
     10 Golden Questions\\
     Pre- and Post-SFT Analysis};

% ════════════════════════════════════════════════════
% SOLID arrow: Dual-Stage Eval → Qualitative Case Study
% Exits eval.west, goes left past metrics, enters qual.west
% ════════════════════════════════════════════════════
\draw[arrow]
    (eval.west)
    -- ++(-7.5, 0)
    -- ++(0, -6.6)
    -| (qual.west);

\end{tikzpicture}
}% end resizebox
\caption{Experimental pipeline: data preparation, parameter-efficient
fine-tuning, and evaluation across three model architectures.}
\label{fig:pipeline}
\end{figure}

This section describes the complete experimental pipeline
consisting of seven sequential stages.
\textbf{Stage~1} covers source corpus selection from an
existing Devanagari instruction-following dataset.
\textbf{Stage~2} applies bilingual transformation: the
first 5,000 samples undergo selective English instruction
translation while retaining Romanized Nepali
\textit{Input} and \textit{Output} fields, and the
remaining 5,000 samples undergo full phonetic
transliteration of all three Alpaca fields to Romanized
Nepali.
\textbf{Stage~3} partitions the transformed 10,000-sample
corpus into a 9,000-sample training set and a 1,000-sample
held-out test set.
\textbf{Stage~4} applies parameter-efficient fine-tuning
via QLoRA~\cite{dettmers2024qlora} with
rsLoRA~\cite{kalajdzievski2023rslora} at rank $r=32$
on dual NVIDIA Tesla T4 GPUs across all three
architectures.
\textbf{Stage~5} conducts dual-stage evaluation comparing
zero-shot and fine-tuned performance on the held-out
test set.
\textbf{Stage~6} scores all outputs across five metrics
spanning seven measurement dimensions: PPL, BERTScore,
chrF++, ROUGE-1, ROUGE-2, ROUGE-L, and BLEU
\textbf{Stage~7} performs qualitative case study analysis
across 10 Golden Questions.
Figure~\ref{fig:pipeline} provides a visual overview of
the full pipeline.

% ── Replace your existing figure block with this ──

\subsection{Dataset Construction and Transliteration}
% ────────────────────────────────────────────────────────────

\subsubsection{Source Corpus Selection}

The primary data source is the
\textit{Saugatkafley/alpaca-nepali-sft}
dataset~\cite{kafley2024alpaca}, an instruction-following
corpus originally in Devanagari script containing
approximately 52,000 samples. We extracted a curated
subset of \textbf{10,000 samples}, maximising diversity
across instruction types including logical reasoning,
creative writing, and factual Q\&A. Sequences exceeding
512 tokens were truncated for computational efficiency.

\subsubsection{Bilingual Transformation Pipeline}

All 10,000 samples originate in Devanagari and undergo
one of two transformation paths depending on their
partition assignment, as illustrated in Stage~2 of
Figure~\ref{fig:pipeline}:

\begin{enumerate}
    \item \textbf{Semantic Translation (5,000 samples).}
    For the first 5,000 samples, only the
    \textbf{Instruction} field was translated from
    Devanagari into English using the Google Translate
    engine~\cite{googletranslate2024}. The corresponding
    \textbf{Input} and \textbf{Output} fields were
    converted to Romanized Nepali via phonetic
    transliteration using the
    \texttt{Indic-Transliteration}
    library~\cite{indictrans2024}. This asymmetric
    design enables the model to learn the mapping
    between high-resource English prompting and
    low-resource Romanized Nepali responses, improving
    cross-lingual instruction following without
    sacrificing output-script consistency.

    \item \textbf{Full Phonetic Transliteration
    (5,000 samples).}
    For the remaining 5,000 samples, all three Alpaca
    fields  \textbf{Instruction}, \textbf{Input},
    and \textbf{Output}  were converted from
    Devanagari to Romanized Nepali using the
    \texttt{Indic-Transliteration}
    library~\cite{indictrans2024}. Phonetic variations
    were intentionally permitted (e.g., mapping
    \textit{`chha'} to both \textit{`cha'} and
    \textit{`chha'}) to replicate the non-standardized
    orthography of real-world digital Nepali
    communication.
\end{enumerate}

The resulting dataset contains a 50/50 split of
English-instructed and Romanized-instructed samples,
with all \textbf{Output} fields consistently in
Romanized Nepali across both partitions. All samples
follow the Alpaca instruction-following
schema~\cite{taori2023alpaca}
(\textit{Instruction, Input, Output}).

\subsubsection{Train/Test Partitioning}

Following bilingual transformation, the full 10,000-sample
corpus was partitioned into a
\textbf{9,000-sample training set} and a
\textbf{1,000-sample held-out test set} (the ``1k test
set''). The test set was sequestered before any training
decision to prevent data leakage and is used exclusively
for evaluation in Stages~5--7 of the pipeline.

% ────────────────────────────────────────────────────────────
\subsection{Parameter-Efficient Fine-Tuning with QLoRA}
\label{sec:qlora}
% ────────────────────────────────────────────────────────────

To adapt all three models within the constraints of dual
NVIDIA Tesla T4 GPUs (16\,GB VRAM each), we employed
QLoRA~\cite{dettmers2024qlora} accelerated via the
\textbf{Unsloth} framework~\cite{unsloth2024}. The
fine-tuning stack combines 4-bit NF4 quantization,
low-rank adapter injection, and rank-stabilized scaling.

\subsubsection{4-bit NormalFloat (NF4) Quantization}

All base model weights were frozen and compressed to
\textbf{4-bit NF4} precision~\cite{dettmers2024qlora}.
NF4 is an information-theoretically optimal 4-bit data
type whose $2^4 = 16$ quantization levels are
non-uniformly spaced to minimise expected quantization
error under a standard normal prior $\mathcal{N}(0,1)$,
unlike uniformly spaced INT4. Quantization is applied
block-wise:

\begin{equation}
    W_{\text{NF4}} = \text{quantize}_{\text{NF4}}\!\left(
        \frac{W}{\max|W|}
    \right)
    \label{eq:nf4}
\end{equation}

Per-block scaling constants are stored in 32-bit float
for dequantization at inference. Base weights receive no
gradient updates; only adapter matrices are trained.

\subsubsection{Low-Rank Adapter Injection}
\label{sec:lora_inject}

Following Hu et al.\ \cite{hu2022lora}, trainable
low-rank matrices are injected alongside each target
weight. For a frozen weight
$W_{\text{NF4}} \in \mathbb{R}^{d \times d}$,
the adapted forward pass becomes:

\begin{equation}
    h = W_{\text{NF4}}\,x + \frac{\alpha}{r}\,B A\,x
    \label{eq:lora_forward}
\end{equation}

where $A \in \mathbb{R}^{r \times d}$ and
$B \in \mathbb{R}^{d \times r}$ are trainable adapter
matrices, $r \ll d$ is the adapter rank, and $\alpha$
is a scaling hyperparameter. At initialisation,
$A \sim \mathcal{N}(0,\sigma^2)$ and $B = \mathbf{0}$,
ensuring zero adapter contribution at step 0. Adapters
were injected into seven projection layers per
transformer block:

\[
    \text{Target modules} = \{
        \texttt{q\_proj},\;
        \texttt{k\_proj},\;
        \texttt{v\_proj},\;
        \texttt{o\_proj},\;
        \texttt{gate\_proj},\;
        \texttt{up\_proj},\;
        \texttt{down\_proj}
    \}
\]

All other weights remained frozen. We selected $r = 32$
and $\alpha = 64$, higher than common defaults of
$r \in \{8, 16\}$~\cite{hu2022lora}, to provide
sufficient capacity for the complex mapping from
English-centric pre-training to Romanized Nepali
phonetics.

\subsubsection{Rank-Stabilized Scaling (rsLoRA)}

Standard LoRA's $\alpha/r$ scaling progressively
suppresses adapter influence as rank
increases~\cite{kalajdzievski2023rslora}. At $r=32$
this is non-trivial. rsLoRA replaces $1/r$ with
$1/\sqrt{r}$:

\begin{equation}
    h = W_{\text{NF4}}\,x
        \;+\; \frac{\alpha}{\sqrt{r}}\,B A\,x
    \label{eq:rslora}
\end{equation}

For our configuration ($r=32$, $\alpha=64$):

\begin{equation}
    \frac{\alpha}{\sqrt{r}} = \frac{64}{\sqrt{32}}
    \approx 11.31
    \quad \text{vs.} \quad
    \frac{\alpha}{r} = \frac{64}{32} = 2.0
\end{equation}

This $5.7\times$ increase in effective adapter
contribution ensures high-rank adapters meaningfully
influence the forward pass without gradient
instability~\cite{kalajdzievski2023rslora}.

% ────────────────────────────────────────────────────────────
\subsection{Experimental Protocol}
% ────────────────────────────────────────────────────────────

The study follows a structured three-phase protocol: (i) Zero-Shot Baseline Assessment to quantify innate cross-lingual transfer, (ii) Supervised Fine-Tuning (SFT) on a curated bilingual Romanized Nepali instruction corpus using QLoRA with rsLoRA, and (iii) Dual-Stage Evaluation comprising quantitative
benchmarking across five metrics spanning seven
measurement dimensions and a qualitative case study
across 10 Golden Questions. All three phases share the same held-out 1,000-sample test set to prevent data leakage, with all reported metrics reflecting best-checkpoint weights recovered via minimum validation loss.

\subsubsection{Phase 1: Zero-Shot Baseline Assessment}

Unmodified base weights of all three models were
evaluated on the 1k test set with no system prompt,
template, or few-shot examples. This zero-shot condition
quantifies each model's innate ability to process
Romanized Nepali arising solely from
pre-training~\cite{brown2020gpt3}, establishing the
\textbf{performance floor} from which all fine-tuning
gains $\Delta$ are measured.

\subsubsection{Phase 2: Supervised Fine-Tuning (SFT)}

All three models were fine-tuned on the 9,000-sample
training partition using the \texttt{SFTTrainer} from
the TRL library~\cite{vonwerra2022trl} in
\textit{Instruction\,--\,Output} format. The QLoRA +
rsLoRA configuration from Section~\ref{sec:qlora} was
applied identically across all architectures so that
Phase~3 differences are attributable to architectural
properties, not configuration asymmetry.

\paragraph{Optimizer and Learning Rate Schedule.}
An 8-bit AdamW optimizer~\cite{dettmers2022adamw8bit}
was used with a peak learning rate of
$1 \times 10^{-4}$. A 200-step linear warmup prevented
gradient divergence during initial exposure to Romanized
text, followed by cosine
decay~\cite{loshchilov2017cosine}.

\paragraph{Batch Configuration.}
A micro-batch size of 2 per device with 4 gradient
accumulation steps yields an effective batch size of:
\begin{equation}
    B_{\text{eff}}
    = B_{\text{device}} \times N_{\text{accum}}
      \times N_{\text{GPU}}
    = 2 \times 4 \times 2 = 16
\end{equation}
All models were trained for 3 epochs, corresponding to
3,375 optimizer steps.

\paragraph{Golden Point Checkpoint Recovery.}
Terminal-epoch overfitting is a documented risk in
low-resource PEFT settings~\cite{dettmers2024qlora}.
Checkpoints and validation evaluations were synchronised
every 100 steps with a rolling window of 10 saved
checkpoints. Upon completion,
\texttt{load\_best\_model\_at\_end} restores the
checkpoint of minimum validation loss
$\mathcal{L}_{\text{eval}}$. Since:
\begin{equation}
    \text{PPL} = \exp\!\left(\mathcal{L}_{\text{eval}}\right)
\end{equation}
minimising $\mathcal{L}_{\text{eval}}$ directly optimises
the primary metric, ensuring all Phase~3 scores reflect
\textit{best-checkpoint} weights.
\begin{table}[h!]
\centering
\small
\begin{tabular}{lll}
\hline
\textbf{Category} & \textbf{Parameter} & \textbf{Value} \\
\hline
\multirow{3}{*}{\textbf{Optimization}}
  & Optimizer
    & AdamW (8-bit)~\cite{dettmers2022adamw8bit} \\
  & Learning Rate        & $1 \times 10^{-4}$ \\
  & Warmup Steps         & 200 \\
\hline
\multirow{2}{*}{\textbf{Batch Logic}}
  & Per-Device Batch Size     & 2 \\
  & Effective Batch Size
    & 16\;($2 \times 4 \times 2$\,GPUs) \\
\hline
\multirow{2}{*}{\textbf{Hardware}}
  & GPUs
    & Dual NVIDIA Tesla T4 (16\,GB each) \\
  & Quantization
    & 4-bit NF4~\cite{dettmers2024qlora} \\
\hline
\multirow{4}{*}{\textbf{LoRA Adapters}}
  & Rank $r$             & 32 \\
  & Alpha $\alpha$       & 64 \\
  & Target Modules
    & \texttt{q, k, v, o\_proj,} \\
  &
    & \texttt{gate, up, down\_proj} \\
  & Scaling
    & rsLoRA\;($\alpha/\sqrt{r}$)~\cite{kalajdzievski2023rslora} \\
\hline
\multirow{5}{*}{\textbf{Checkpointing}}
  & \texttt{save\_strategy}   & \texttt{steps} \\
  & \texttt{save\_steps}      & 100 \\
  & \texttt{save\_total\_limit} & 10 \\
  & \texttt{load\_best\_model\_at\_end} & \texttt{True} \\
  & \texttt{metric\_for\_best\_model}  & \texttt{eval\_loss} \\
\hline
\end{tabular}
\caption{Training, hardware, adapter, and checkpointing
configuration applied identically to all three models.}
\label{tab:training_configs}
\end{table}
\subsubsection{Phase 3: Dual-Stage Evaluation}

\paragraph{Stage A:Quantitative Benchmarking.}
All seven measurement dimensions across five metrics
(PPL, BERTScore, chrF++, ROUGE-1, ROUGE-2, ROUGE-L,
BLEU) were computed on the 1k test set for both zero-shot
(Phase~1) and fine-tuned (Phase~2) weights. For each
model $m$ and metric $\mathcal{M}$, the absolute
fine-tuning gain is:
\begin{equation}
    \Delta\mathcal{M}_{m}
    = \mathcal{M}_{m}^{\text{fine-tuned}}
    - \mathcal{M}_{m}^{\text{zero-shot}}
\end{equation}

\paragraph{Stage B:Qualitative Case Study.}
A manual analysis of 10 \textit{Golden Questions}
spanning factual Q\&A, creative writing, logical
reasoning, and conversational response was conducted
pre- and post-SFT. Full outputs are in
Appendix~\ref{app:golden_questions};
analysis is in Section~\ref{sec:qualitative}.

\begin{table}[H]
\centering
\small
\begin{tabular}{p{1.0cm} p{2.0cm} p{4.0cm} p{4.0cm}}
\hline
\textbf{Phase} & \textbf{Name} & \textbf{Method}
              & \textbf{Output} \\
\hline
1 &
Zero-Shot &
Unmodified weights; raw Romanized input; no prompting &Performance floor across five metrics,
seven measurement dimensions \\
\hline
2 &
SFT &
QLoRA~\cite{dettmers2024qlora} +
rsLoRA~\cite{kalajdzievski2023rslora};
4-bit NF4; 3 epochs; Golden Point recovery &
Best-checkpoint adapters per model \\
\hline
3A &
Quantitative &
5-metric 7 dimension scoring; zero-shot vs.\ fine-tuned;
$\Delta$ per metric &
Per-model gain table \\
\hline
3B &
Qualitative &
10 Golden Questions; pre- and post-SFT &
Linguistic failure analysis \\
\hline
\end{tabular}
\caption{Summary of the three experimental phases.}
\label{tab:phase_summary}
\end{table}

% ────────────────────────────────────────────────────────────
% ────────────────────────────────────────────────────────────
\subsection{Evaluation Metrics}
% ────────────────────────────────────────────────────────────

% CHANGE TO:
Five complementary metrics spanning seven measurement
dimensions cover fluency, surface similarity, and
semantic alignment. ROUGE is reported across three
variants (ROUGE-1, ROUGE-2, ROUGE-L) to capture
unigram, bigram, and longest common subsequence
alignment respectively. Together they
capture orthographic, structural, and semantic
dimensions of generation quality  a necessary
multi-axis framework given the non-standardized
orthography of Romanized Nepali, where surface-form
variation is high but semantic intent is often
preserved~\cite{shahi2022review}.

% ── PPL ─────────────────────────────────────────────────────
\subsubsection{Perplexity (PPL)}

PPL measures intrinsic language model
fluency~\cite{jurafsky2023speech}. For a tokenized
sequence $X = (x_1, \dots, x_t)$:

\begin{equation}
    \text{PPL}(X) = \exp\!\left(
        -\frac{1}{t} \sum_{i=1}^{t}
        \log P_{\theta}(x_i \mid x_{<i})
    \right)
    \label{eq:ppl}
\end{equation}

Lower PPL indicates greater predictive confidence.
PPL is the primary metric optimised by the Golden
Point checkpoint strategy, since
$\text{PPL} = \exp(\mathcal{L}_{\text{eval}})$,
where $\mathcal{L}_{\text{eval}}$ is the average
negative log-likelihood over the held-out test set.
Minimising $\mathcal{L}_{\text{eval}}$ therefore
directly minimises PPL.

% ── chrF++ ──────────────────────────────────────────────────
\subsubsection{chrF++}

chrF++~\cite{popovic2015chrf} computes a character
$n$-gram F-score. It is preferred over word-level
metrics such as BLEU for transliterated scripts
because character $n$-grams are robust to the
orthographic variation inherent in Romanized Nepali,
where the same word may be spelled in multiple
phonetically equivalent ways~\cite{shahi2022review}:

\begin{equation}
    \text{chrF}_{\beta} =
    \frac{(1 + \beta^2)\,\text{Prec}\cdot\text{Rec}}%
         {\beta^2 \cdot \text{Prec} + \text{Rec}}
    \label{eq:chrf}
\end{equation}

We set $\beta = 2$ to weight recall more heavily,
as missing Nepali morphological suffixes are more
costly than spurious ones.

% ── BERTScore ───────────────────────────────────────────────
\subsubsection{BERTScore}

BERTScore~\cite{zhang2019bertscore} evaluates
semantic similarity via contextual embeddings,
capturing meaning beyond surface token matching.
This makes it particularly valuable for Romanized
Nepali, where orthographic variants of the same
word should receive high similarity scores even
when exact string matching fails. For reference
$r$ and candidate $c$ with unit-normalized
contextual embeddings $\{x_i\}$ and $\{y_j\}$,
the recall component is:

\begin{equation}
    R_{\text{BERT}} = \frac{1}{|r|}
        \sum_{x_i \in r}
        \max_{y_j \in c}\;
        \cos(x_i,\, y_j)
    \label{eq:bertscore}
\end{equation}

where $\cos(x_i, y_j) = x_i^{\top} y_j /
(\|x_i\| \|y_j\|)$ is the cosine similarity
between embedding pairs. The final $F_1$ score
combines precision and recall analogously.

% ── ROUGE-L ─────────────────────────────────────────────────
\subsubsection{ROUGE-L}

ROUGE-L~\cite{lin2004rouge} assesses structural
alignment via the Longest Common Subsequence (LCS),
capturing fluency and word-order preservation without
requiring contiguous $n$-gram matches:

\begin{equation}
    F_{\text{LCS}} =
    \frac{(1 + \beta^2)\,
        P_{\text{LCS}}\, R_{\text{LCS}}}%
        {\beta^2\, P_{\text{LCS}} + R_{\text{LCS}}}
    \label{eq:rouge}
\end{equation}

ROUGE-1 (unigram overlap) and ROUGE-2 (bigram overlap)
are reported alongside ROUGE-L, together constituting
three of the seven measurement dimensions and providing
a complete picture of structural alignment at multiple
granularities.

% ── BLEU ────────────────────────────────────────────────────
\subsubsection{BLEU}

BLEU~\cite{papineni2002bleu} is the standard 4-gram
precision metric included for compatibility with
prior NLP literature:

\begin{equation}
    \text{BLEU} = BP \cdot \exp\!\left(
        \sum_{n=1}^{4} w_n \log p_n
    \right), \quad w_n = \tfrac{1}{4}
    \label{eq:bleu}
\end{equation}

where $BP$ is the brevity penalty and $p_n$ is
the $n$-gram precision for order $n$.
Given the orthographic variance of Romanized
Nepali, BLEU is expected to underestimate true
generation quality~\cite{post2018call} because
it relies on exact surface-form matches. It is
therefore interpreted alongside chrF++ and
BERTScore~\cite{popovic2015chrf, zhang2019bertscore},
which are more robust to spelling variation.
% ════════════════════════════════════════════════════════════
\clearpage
\section{Results}
% ════════════════════════════════════════════════════════════

% ────────────────────────────────────────────────────────────
\subsection{Zero-Shot Baseline Performance}
% ────────────────────────────────────────────────────────────

Table~\ref{tab:zero_shot} reports baseline performance.
Qwen3-8B and Mistral-7B achieve comparable perplexity
(27.89 and 27.53 respectively), both substantially
outperforming Llama-3.1-8B ($\text{PPL} = 52.79$),
suggesting weaker innate cross-lingual
transfer~\cite{pires2019multilingual}. Qwen3-8B leads on
semantic alignment with the highest BERTScore (0.5631) and
chrF++ (13.10), consistent with its broader multilingual
pre-training~\cite{bai2023qwen}. Mistral-7B records the
lowest BERTScore (0.2327) and chrF++ (4.95), indicating
it models token-level distributions competently but fails
to produce semantically coherent output without adaptation.

\begin{table}[h!]
\centering
\small
\begin{tabular}{lrrrrrrr}
\hline
\textbf{Model} &
\textbf{PPL}$\downarrow$ & \textbf{BERT}$\uparrow$ &
\textbf{BLEU}$\uparrow$ & \textbf{chrF++}$\uparrow$ &
\textbf{R-1}$\uparrow$ & \textbf{R-2}$\uparrow$ &
\textbf{R-L}$\uparrow$ \\
\hline
Llama-3.1-8B
  & 52.79 & 0.4224 & 0.0143 & 8.15
  & 0.0732 & 0.0311 & 0.0681 \\
Qwen3-8B
  & 27.89 & 0.5631 & 0.0099 & 13.10
  & 0.0570 & 0.0195 & 0.0513 \\
Mistral-7B
  & 27.53 & 0.2327 & 0.0105 & 4.95
  & 0.0309 & 0.0134 & 0.0281 \\
\hline
\end{tabular}
\caption{Zero-shot baseline performance on the 1k test
set. $\downarrow$ = lower is better;
$\uparrow$ = higher is better.}
\label{tab:zero_shot}
\end{table}

% ────────────────────────────────────────────────────────────
\subsection{Fine-Tuned Performance}
% ────────────────────────────────────────────────────────────

Following three epochs of QLoRA + rsLoRA SFT with Golden
Point recovery, all models converge to perplexity
$2.81$--$3.02$ (Table~\ref{tab:finetuned}). Qwen3-8B
achieves the highest chrF++ (27.47) and ROUGE scores.
Llama-3.1-8B records the highest BERTScore
(0.7511)~\cite{zhang2019bertscore}. Mistral-7B achieves
the lowest perplexity (2.812) but trails on surface metrics.

\begin{table}[h!]
\centering
\small
\begin{tabular}{lrrrrrrr}
\hline
\textbf{Model} &
\textbf{PPL}$\downarrow$ & \textbf{BERT}$\uparrow$ &
\textbf{BLEU}$\uparrow$ & \textbf{chrF++}$\uparrow$ &
\textbf{R-1}$\uparrow$ & \textbf{R-2}$\uparrow$ &
\textbf{R-L}$\uparrow$ \\
\hline
Llama-3.1-8B
  & 3.024 & \textbf{0.7511} & 0.0498 & 26.97
  & 0.2756 & 0.1002 & 0.2359 \\
Qwen3-8B
  & 2.946 & 0.7505 & \textbf{0.0550} & \textbf{27.47}
  & \textbf{0.2915} & \textbf{0.1162} & \textbf{0.2511} \\
Mistral-7B
  & \textbf{2.812} & 0.7339 & 0.0404 & 23.95
  & 0.2460 & 0.0842 & 0.2144 \\
\hline
\end{tabular}
\caption{Fine-tuned performance after QLoRA + rsLoRA SFT
with Golden Point recovery. Bold = best per metric.}
\label{tab:finetuned}
\end{table}

% ────────────────────────────────────────────────────────────

\subsection{Fine-Tuning Gain Analysis}
% ────────────────────────────────────────────────────────────

Table~\ref{tab:delta} reports the absolute fine-tuning
gain ($\Delta$ = fine-tuned $-$ zero-shot) for each
model across five metrics spanning seven measurement dimensions. Llama-3.1-8B records
the largest PPL reduction ($-49.77$) despite its
weakest zero-shot baseline, confirming the adaptation
headroom hypothesis~\cite{hu2022lora, dettmers2024qlora}.
Mistral-7B-v0.1 achieves the largest BERTScore gain
($+0.5012$) and chrF++ gain ($+19.00$) from its lowest
zero-shot semantic baseline  a direct consequence of
its near-zero semantic starting point amplifying
absolute gain magnitude. Qwen3-8B leads on all three
structural alignment metrics (ROUGE-1: $+0.2345$,
ROUGE-2: $+0.0967$, ROUGE-L: $+0.1998$) and BLEU
($+0.0451$), owing to its stronger multilingual
pre-training coverage~\cite{bai2023qwen}.

\begin{table}[H]
\centering
\small
\setlength{\tabcolsep}{4.5pt}
\begin{tabular}{lrrrrrrr}
\hline
\textbf{Model} &
$\Delta$\textbf{PPL}$\downarrow$ &
$\Delta$\textbf{BERT}$\uparrow$ &
$\Delta$\textbf{BLEU}$\uparrow$ &
$\Delta$\textbf{chrF++}$\uparrow$ &
$\Delta$\textbf{R-1}$\uparrow$ &
$\Delta$\textbf{R-2}$\uparrow$ &
$\Delta$\textbf{R-L}$\uparrow$ \\
\hline
Llama-3.1-8B
  & \textbf{$-$49.77}
  & $+$0.3287
  & $+$0.0355
  & $+$18.82
  & $+$0.2024
  & $+$0.0691
  & $+$0.1678 \\
Qwen3-8B
  & $-$24.95
  & $+$0.1874
  & \textbf{$+$0.0451}
  & $+$14.37
  & \textbf{$+$0.2345}
  & \textbf{$+$0.0967}
  & \textbf{$+$0.1998} \\
Mistral-7B
  & $-$24.72
  & \textbf{$+$0.5012}
  & $+$0.0299
  & \textbf{$+$19.00}
  & $+$0.2151
  & $+$0.0708
  & $+$0.1863 \\
\hline
\multicolumn{8}{l}{\small\textit{Bold = largest
gain per column.\;
$\downarrow$ = lower is better;\;
$\uparrow$ = higher is better.}}\\
\hline
\end{tabular}
\caption{Absolute fine-tuning gain ($\Delta$ =
fine-tuned $-$ zero-shot) across five metrics spanning seven measurement dimensions.}
\label{tab:delta}
\end{table}
% ────────────────────────────────────────────────────────────
% ────────────────────────────────────────────────────────────
\subsection{Training and Validation Loss Analysis}
% ────────────────────────────────────────────────────────────

Training and validation loss curves for all three models
over 3 epochs (3,375 steps) are presented individually
below. Checkpoints were saved every 100 steps and the
best checkpoint was recovered using
\texttt{load\_best\_model\_at\_end} based on minimum
validation loss~\cite{dettmers2024qlora}.

% ── Llama ───────────────────────────────────────────────────
\begin{figure}[H]
    \centering
    \includegraphics[width=\textwidth]{llama_loss.png}
    \caption{Training and validation loss for
    Llama-3.1-8B over 3,375 steps. The green dotted
    line marks the best checkpoint at step~2,200
    (minimum validation loss = 1.1585).}
    \label{fig:llama_loss}
\end{figure}

Llama-3.1-8B exhibits the highest initial training loss
($\approx 2.10$) among all three models, consistent with
its Tiktoken tokenizer over-fragmenting Romanized Nepali
into low-frequency subword units~\cite{rust2021good}.
Validation loss decreases steadily through Epochs~1 and~2,
reaching a global minimum of \textbf{1.1585} at
step~2,200 (end of Epoch~2). At the Epoch~3 boundary
(step~2,200--2,300) a sharp spike in validation loss to
$\approx 1.28$ is observed, caused by the learning rate
schedule resetting. The Golden Point recovery mechanism
correctly identifies step~2,200 as the optimal checkpoint,
preventing the Epoch~3 degradation from affecting the
final reported metrics.

% ── Mistral ─────────────────────────────────────────────────
\begin{figure}[H]
    \centering
    \includegraphics[width=\textwidth]{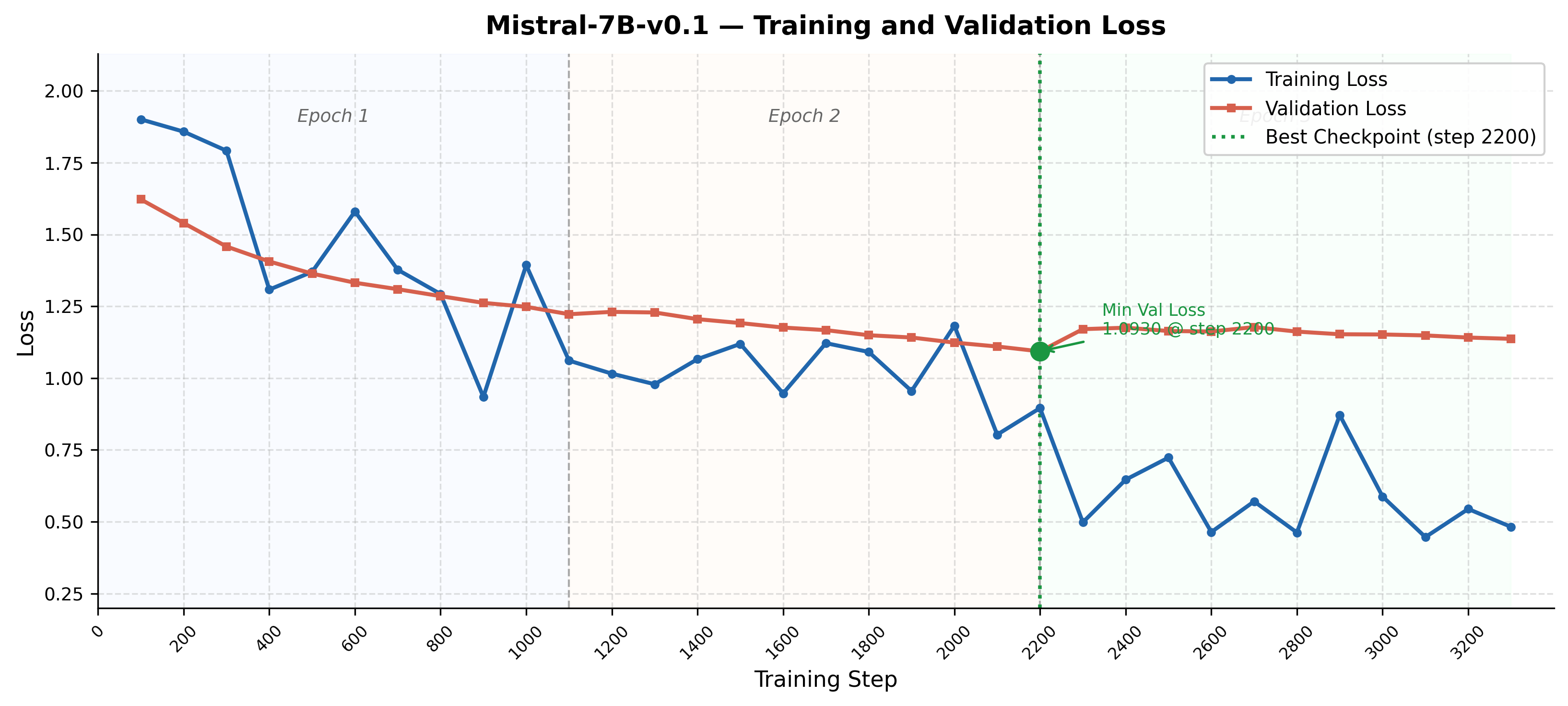}
    \caption{Training and validation loss for
    Mistral-7B-v0.1 over 3,375 steps. The green dotted
    line marks the best checkpoint at step~2,200
    (minimum validation loss = 1.0930).}
    \label{fig:mistral_loss}
\end{figure}

Mistral-7B-v0.1 begins with the lowest initial training
loss ($\approx 1.90$) of the three models, reflecting its
SentencePiece tokenizer producing more coherent subword
units for Latin-script input than Tiktoken. The validation
curve is the smoothest of all three architectures,
declining consistently across both Epochs~1 and~2 with
minimal oscillation. The global validation minimum of
\textbf{1.0930} is achieved at step~2,200, the lowest
among all three models. An epoch-boundary spike to
$\approx 1.17$ is visible at step~2,300 before Epoch~3
stabilizes, again confirming that Golden Point recovery
at step~2,200 yields the best generalization state.

% ── Qwen ────────────────────────────────────────────────────
\begin{figure}[H]
    \centering
    \includegraphics[width=\textwidth]{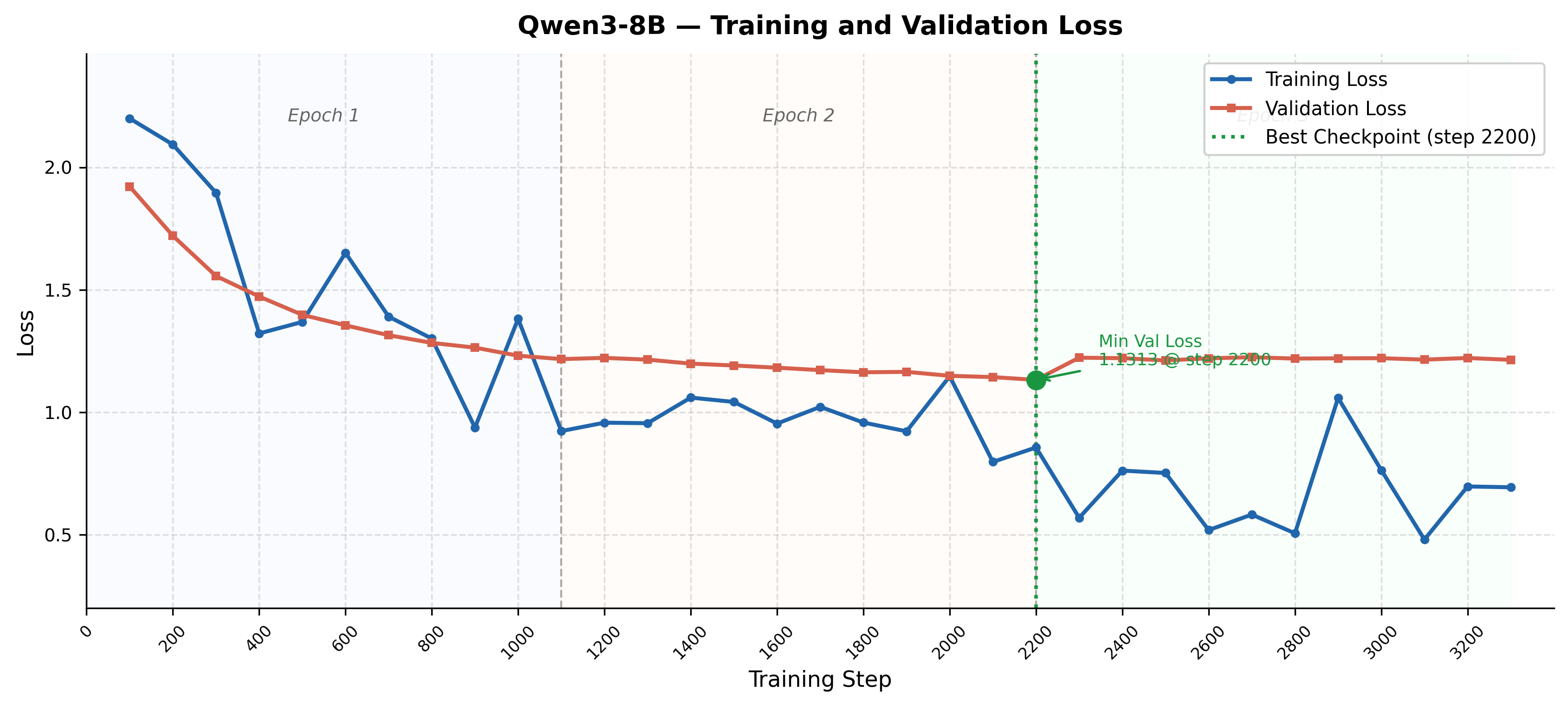}
    \caption{Training and validation loss for Qwen3-8B
    over 3,375 steps. The green dotted line marks the
    best checkpoint at step~2,200 (minimum validation
    loss = 1.1313).}
    \label{fig:qwen_loss}
\end{figure}

Qwen3-8B starts with an intermediate initial training
loss ($\approx 2.20$) and converges steadily through
Epochs~1 and~2. The validation loss reaches its global
minimum of \textbf{1.1313} at step~2,200. Notably,
Qwen3-8B's Epoch~3 validation loss plateaus at
$\approx 1.22$ rather than recovering toward the
Epoch~2 minimum, suggesting the adapter reached a
capacity ceiling earlier than the other two models
consistent with Qwen3-8B's stronger multilingual
pre-training baseline reducing available adaptation
headroom~\cite{bai2023qwen}. The Epoch~3 spike is
present but less pronounced than in Llama-3.1-8B.

% ── Cross-model summary ─────────────────────────────────────
\paragraph{Cross-Model Summary.}
All three models achieve their global validation minimum
at \textbf{step~2,200}, empirically confirming that
Golden Point checkpoint recovery was necessary and
effective across all architectures. Mistral-7B-v0.1
achieves the lowest absolute validation loss (1.0930),
followed by Qwen3-8B (1.1313) and Llama-3.1-8B (1.1585).
However, as shown in Tables~\ref{tab:finetuned}
and~\ref{tab:delta}, lower validation loss does not
directly predict superior downstream metric performance
 Llama-3.1-8B achieves the highest BERTScore (0.7511)
despite its highest validation loss, supporting the
adaptation headroom hypothesis discussed in
Section~\ref{sec:qualitative}.

% ────────────────────────────────────────────────────────────
\subsection{Trainable Parameters and Training Cost}
% ────────────────────────────────────────────────────────────

\begin{table}[H]
\centering
\small
\begin{tabular}{lrrr}
\hline
\textbf{Property} &
\textbf{Llama-3.1-8B} & \textbf{Qwen3-8B} &
\textbf{Mistral-7B-v0.1} \\
\hline
\multicolumn{4}{l}{\textit{Model Scale}} \\
\hline
Total Parameters
  & 8,114,147,328 & 8,278,029,312 & 7,325,618,176 \\
Trainable Parameters
  & 83,886,080    & 87,293,952    & 83,886,080    \\
Trainable \%
  & 1.03\%        & 1.05\%        & 1.15\%        \\
\hline
\multicolumn{4}{l}{\textit{Training Cost}} \\
\hline
Wall-Clock Time
  & 8h 50m 51s & 8h 26m 01s & 9h 09m 31s \\
Total GPU Time
  & \multicolumn{3}{r}{26h 26m 23s} \\
\hline
\end{tabular}
\caption{Trainable parameter counts and wall-clock
training time from Unsloth~\cite{unsloth2024} runtime
logs. All models: 3 epochs / 3,375 steps, rank
$r=32$~\cite{hu2022lora}, dual NVIDIA Tesla T4.}
\label{tab:training_cost}
\end{table}

% ────────────────────────────────────────────────────────────
\clearpage
\subsection{Qualitative Case Study}
\label{sec:qualitative}
% ────────────────────────────────────────────────────────────

Full outputs for all 10 Golden Questions are in
Appendix~\ref{app:golden_questions}
(Tables~\ref{tab:llama_base}--\ref{tab:mistral_ft}).

\paragraph{Zero-Shot Failure Signatures.}
Three distinct architecture-specific failures
emerge~\cite{shahi2022review}. Llama-3.1-8B produced
\textit{[no response]} on 6 of 10 instructions,
indicating Tiktoken over-fragmented Romanized Nepali into
low-frequency subword units~\cite{rust2021good}.
Mistral-7B-v0.1 produced \textit{[empty line]} on 5 of
10 and responded off-task in English on the remaining 5.
Qwen3-8B was the only model to produce non-empty output
on all 10, yet exclusively in English or Devanagari,
reflecting misdirected cross-lingual
transfer~\cite{pires2019multilingual}.

\paragraph{Post-SFT Resolution.}
Following fine-tuning~\cite{dettmers2024qlora}, all
three models produced Romanized Nepali on all 10
instructions. Llama-3.1-8B and Qwen3-8B generated
fluent, semantically complete responses. Mistral-7B-v0.1
retained two residual semantic errors, consistent with
its lower chrF++ (23.95) and ROUGE-L
(0.2144)~\cite{popovic2015chrf, lin2004rouge}.

\paragraph{Failure Taxonomy Summary.}
For \textbf{Orthographic Robustness}, all fine-tuned
models correctly resolved phonetically variant inputs.
For \textbf{Morphological Integrity}, Llama-3.1-8B and
Qwen3-8B correctly generated Nepali verb endings and
postpositions, while Mistral occasionally dropped them.
For \textbf{Latent Romanization Drift}, all fine-tuned
models maintained Romanized Nepali without drifting into
English syntax~\cite{pires2019multilingual}.

% ════════════════════════════════════════════════════════════
\section{Discussion}
% ════════════════════════════════════════════════════════════

\subsection{Base Model Structural Collapse}

All three base models were functionally unable to respond
in Romanized Nepali at zero-shot
(Table~\ref{tab:failure_analysis}), each exhibiting a
distinct failure mode rooted in its architecture and
tokenizer design~\cite{rust2021good, shahi2022review}.
These failures confirm that Romanized Nepali falls outside
the confident generation distribution of all three
pre-training regimes, consistent with the ``linguistic
shadow'' framing of Bender et al.\
\cite{bender2021dangers}.

\begin{table}[H]
\centering
\small
\begin{tabular}{lp{3.0cm}p{7.0cm}}
\hline
\textbf{Model} & \textbf{Failure Mode}
              & \textbf{Technical Description} \\
\hline
Llama-3.1-8B
  & Semantic Void
  & \textbf{Early EOS Triggering}~\cite{touvron2023llama}:
    Tiktoken over-fragments Romanized Nepali; model
    immediately predicts End-of-Sequence, yielding null
    output on 6 of 10 instructions. \\
\hline
Mistral-7B-v0.1
  & Newline Loop
  & \textbf{Distributional
    Collapse}~\cite{jiang2023mistral}: SentencePiece
    produces plausible fragments but the model fails to
    initiate a response, defaulting to continuous
    \texttt{\textbackslash n} generation. \\
\hline
Qwen3-8B
  & Script Drift
  & \textbf{Latent Script
    Bias}~\cite{bai2023qwen}: Model correctly identifies
    semantic intent but defaults to dominant pre-training
    scripts rather than Romanized format. \\
\hline
\end{tabular}
\caption{Taxonomy of structural failures in zero-shot
base models.}
\label{tab:failure_analysis}
\end{table}

\subsection{The Adaptation Headroom Hypothesis}

Llama-3.1-8B presents a compelling inversion of the
expected performance ordering: despite recording the
weakest zero-shot baseline across all seven measurement
dimensions (PPL~52.79, BERTScore~0.4224, chrF++~8.15),
it achieves the largest absolute PPL reduction
($\Delta = -49.77$), the highest post-SFT BERTScore
(0.7511), and the second-largest chrF++ gain
($\Delta = +18.82$) following fine-tuning. This pattern
is not incidental but structural.

We term this the \textit{adaptation headroom hypothesis}:
models with weaker innate cross-lingual representations
possess greater representational plasticity for
script-domain adaptation when exposed to high-quality
PEFT data. The intuition is straightforward a model
that has formed no confident prior over Romanized Nepali
phoneme sequences has more parameter space available for
the adapter to reshape, whereas a model with a strong
multilingual prior (such as Qwen3-8B) approaches a
representational ceiling sooner, as evidenced by its
Epoch~3 validation loss plateauing at $\approx 1.22$
rather than recovering toward its Epoch~2
minimum~\cite{bai2023qwen}.

Formally, for a model $m$ with zero-shot performance
floor $\mathcal{M}_{m}^{\text{zero-shot}}$ and post-SFT
ceiling $\mathcal{M}_{m}^{\text{fine-tuned}}$, the
adaptation headroom $\mathcal{H}_{m}$ on metric
$\mathcal{M}$ is:

\begin{equation}
    \mathcal{H}_{m}(\mathcal{M})
    = \mathcal{M}_{m}^{\text{fine-tuned}}
    - \mathcal{M}_{m}^{\text{zero-shot}}
    \label{eq:headroom}
\end{equation}

Under this definition, Llama-3.1-8B maximises
$\mathcal{H}_{m}(\text{PPL})$ by a factor of $2.0\times$
over both Qwen3-8B ($\Delta = -24.95$) and
Mistral-7B-v0.1 ($\Delta = -24.72$), confirming that
the weakest pre-trained baseline produces the steepest
adaptation trajectory under identical QLoRA +
rsLoRA conditions~\cite{hu2022lora, dettmers2024qlora,
kalajdzievski2023rslora}.

This result carries a non-obvious practical implication:
a strong zero-shot multilingual baseline does not
guarantee the highest fine-tuning return. When the
deployment scenario involves a high-quality curated
corpus and iterative fine-tuning rounds, adaptation
headroom is a more actionable model selection criterion
than zero-shot performance. Practitioners who select
models solely on zero-shot benchmarks risk
systematically overlooking architectures with the
highest long-term fine-tuning yield.
\subsection{Limitations}

Several limitations should be acknowledged. First, the
\texttt{Indic-Transliteration} pipeline approximates but
does not fully reproduce the orthographic diversity of
naturally occurring user-generated Romanized
Nepali, potentially
underestimating real-world phonetic noise. Second,
evaluation is on a single instruction-following
dataset~\cite{kafley2024alpaca}; performance on social
media text, customer service conversations, or
domain-specific content is not assessed. Third,
BERT Score uses a multilingual BERT model with limited
Romanized Nepali exposure, potentially underestimating
semantic similarity for high-quality outputs that diverge
in surface form~\cite{zhang2019bertscore}. Fourth, all
metrics reflect quantized (4-bit NF4) inference, which
introduces a slight precision-performance trade-off whose
magnitude on the reported values has not been
independently quantified~\cite{dettmers2024qlora}.

\subsection{Broader Impact and Applications}

Romanized Nepali is the de facto language of Nepali
social media, messaging platforms, and youth
discourse.
Current NLP infrastructure, being Devanagari-centric, is
effectively inaccessible to this register, creating
barriers for content moderation, mental health chatbots,
educational tools, and diaspora communication services
for the approximately 3 million Nepali speakers
abroad~\cite{census2021nepal}.

By demonstrating that parameter-efficient adaptation of
comparable-sized models achieves BERTScore $\approx 0.75$ with
under 90M trainable parameters and 26 GPU-hours on dual
T4 hardware, this work establishes a practically
replicable path for low-resource NLP practitioners and
Nepali developer communities to build Romanized Nepali
AI tools without access to large compute
clusters~\cite{hu2022lora, dettmers2024qlora}.
% ────────────────────────────────────────────────────────────
\subsection{Overall Model Assessment}
\label{sec:winner}
% ────────────────────────────────────────────────────────────

Drawing on the fine-tuned performance (Table~\ref{tab:finetuned}),
absolute gains across five metrics spanning seven measurement
dimensions (Table~\ref{tab:delta}), training cost
(Table~\ref{tab:training_cost}), and the qualitative case study
(Section~\ref{sec:qualitative}), we consolidate a dimension-wise
post-SFT ranking across all three architectures in
Table~\ref{tab:winner_summary}. Zero-shot baselines are reported
separately in Table~\ref{tab:zero_shot} and are not repeated here.
 
\begin{table}[H]
\centering
\small
\setlength{\tabcolsep}{4pt}
\begin{tabular}{p{3.8cm}p{2.8cm}p{2.8cm}p{2.8cm}p{1.6cm}}
\hline
\textbf{Dimension} &
\textbf{Llama-3.1-8B} &
\textbf{Qwen3-8B} &
\textbf{Mistral-7B-v0.1} &
\textbf{Best} \\
\hline

\multicolumn{5}{l}{\textit{Post-SFT Quality (five metrics, seven dimensions)}} \\
\hline

Post-SFT PPL$\downarrow$
  & 3rd (3.024)
  & 2nd (2.946)
  & \textbf{1st} (2.812)
  & Mistral \\

Post-SFT BERTScore$\uparrow$
  & \textbf{1st} (0.7511)
  & 2nd (0.7505)
  & 3rd (0.7339)
  & Llama \\

Post-SFT chrF++$\uparrow$
  & 2nd (26.97)
  & \textbf{1st} (27.47)
  & 3rd (23.95)
  & Qwen3-8B \\

Post-SFT ROUGE-1$\uparrow$
  & 2nd (0.2756)
  & \textbf{1st} (0.2915)
  & 3rd (0.2460)
  & Qwen3-8B \\

Post-SFT ROUGE-2$\uparrow$
  & 2nd (0.1002)
  & \textbf{1st} (0.1162)
  & 3rd (0.0842)
  & Qwen3-8B \\

Post-SFT ROUGE-L$\uparrow$
  & 2nd (0.2359)
  & \textbf{1st} (0.2511)
  & 3rd (0.2144)
  & Qwen3-8B \\

Post-SFT BLEU$\uparrow$
  & 2nd (0.0498)
  & \textbf{1st} (0.0550)
  & 3rd (0.0404)
  & Qwen3-8B \\

\hline
\multicolumn{5}{l}{\textit{Adaptation \& Cost}} \\
\hline

Adaptation Gain ($\Delta$PPL)$\downarrow$
  & \textbf{1st} ($-$49.77)
  & 2nd ($-$24.95)
  & 3rd ($-$24.72)
  & Llama \\

Training Speed
  & 2nd (8h 51m)
  & \textbf{1st} (8h 26m)
  & 3rd (9h 10m)
  & Qwen3-8B \\

\hline
\multicolumn{5}{l}{\textit{Qualitative}} \\
\hline

Residual Errors (10 Golden Qs)
  & 2nd (2 content)
  & \textbf{1st} (0 errors)
  & 3rd (2 semantic)
  & Qwen3-8B \\

\hline
\multicolumn{4}{l}{\textbf{Dimension wins}} &
\textbf{Qwen: 7 \quad Llama: 2 \quad Mistral: 1} \\

\end{tabular}
\caption{Dimension-wise post-SFT ranking across ten evaluation criteria. Bold indicates best per row.}
\label{tab:winner_summary}
\end{table}
\paragraph{Recommended Architecture: Qwen3-8B.}
Qwen3-8B is the recommended architecture for Romanized
Nepali deployment across the widest range of practical
settings. Post-SFT, it leads all seven measurement
dimensions spanning structural alignment and fluency:
chrF++ (27.47), ROUGE-1 (0.2915), ROUGE-2 (0.1162),
ROUGE-L (0.2511), and BLEU (0.0550)~\cite{bai2023qwen}.
Its qualitative case study produced zero residual errors
against two content errors in Llama-3.1-8B and two
semantic errors in Mistral-7B-v0.1, demonstrating
superior factual accuracy and topic binding. It also
converges fastest at 8h~26m on dual NVIDIA Tesla T4
hardware, making it the most resource-efficient choice
among the three architectures (Table~\ref{tab:training_cost}).

\paragraph{Scenario Exception: Iterative Fine-Tuning.}
Llama-3.1-8B is the preferred architecture when the
development pipeline involves iterative data collection
and repeated fine-tuning rounds. Despite its weakest
zero-shot baseline (PPL~52.79), it achieves the largest
absolute PPL reduction ($\Delta\text{PPL} = -49.77$)
and the highest post-SFT BERTScore (0.7511), confirming
the \textit{adaptation headroom hypothesis}: a weaker
pre-trained multilingual baseline translates into greater
representational plasticity under
PEFT~\cite{hu2022lora, dettmers2024qlora}. Practitioners
building toward a production system over multiple
training iterations on a growing Romanized Nepali corpus
should consider Llama-3.1-8B the more tractable
long-term investment.

\paragraph{Fluency-Only Deployment.}
Mistral-7B-v0.1 is recommended only for applications
where grammatical fluency is the primary requirement
and factual precision is secondary. It achieves the
lowest post-SFT perplexity (2.812) and the largest
BERTScore gain ($\Delta\text{BERTScore} = +0.5012$)
from its near-zero semantic zero-shot baseline, as well
as the smoothest validation loss curve across all three
architectures (minimum $\mathcal{L}_{\text{eval}} =
1.0930$ at step~2{,}200). However, it produced the two
most severe qualitative failures post-SFT: a semantic
hallucination on the Dashain festival task and a
single-token non-answer on the categorical reasoning
task~\cite{jiang2023mistral}. Its lower total parameter
count (7.3B vs.\ 8.1--8.3B) offers a marginal inference
speed advantage on memory-constrained hardware where
perplexity is the dominant deployment criterion.
% ════════════════════════════════════════════════════════════
% ════════════════════════════════════════════════════════════
\section{Conclusion}
% ════════════════════════════════════════════════════════════
% ════════════════════════════════════════════════════════════

% ════════════════════════════════════════════════════════════

This study presents the first systematic benchmarking of
comparable-sized open-weight LLMs on Romanized Nepali, addressing
a persistent and consequential gap in low-resource NLP for
South Asian informal digital communication. We evaluated
Llama-3.1-8B~\cite{touvron2023llama},
Mistral-7B-v0.1~\cite{jiang2023mistral}, and
Qwen3-8B~\cite{bai2023qwen} across zero-shot and fine-tuned
settings under a rigorous framework spanning five metrics across
seven measurement dimensions (PPL, BERTScore, chrF++, ROUGE-1,
ROUGE-2, ROUGE-L, and BLEU), supplemented by an in-depth
qualitative case study across 10 Golden Questions spanning
factual, technical, grammatical, and creative generation tasks.

Four principal findings emerge from this work. First,
all three base models are functionally unable to respond
in Romanized Nepali at zero-shot, each exhibiting a
distinct architecture-specific failure mode: early EOS
triggering in Llama-3.1-8B, distributional collapse in
Mistral-7B-v0.1, and latent script bias in Qwen3-8B.
These failures collectively confirm that Romanized Nepali
lies outside the confident generation distribution of all
comparable-sized model pre-training regimes, consistent with
the ``linguistic shadow'' characterisation of under-resourced
transliterated scripts~\cite{bender2021dangers, shahi2022review}.

Second, QLoRA~\cite{dettmers2024qlora} combined with
rsLoRA~\cite{kalajdzievski2023rslora} at rank $r=32$
fully resolves all three failure modes within 3 epochs
on 9,000 samples, training only $\approx 1\%$ of each
model's parameters on dual NVIDIA Tesla T4 GPUs in under
26 total GPU-hours. Post-SFT, all three architectures
achieve BERTScore $\approx 0.75$ and chrF++ $> 23$,
demonstrating that parameter-efficient fine-tuning is
both necessary and sufficient for script-domain
adaptation at this scale.

Third, the \textit{adaptation headroom hypothesis} is
confirmed: Llama-3.1-8B, despite its weakest zero-shot
baseline (PPL~52.79), achieves the largest absolute PPL
reduction ($\Delta = -49.77$) and the highest post-SFT
BERTScore (0.7511), establishing that a weak pre-trained
multilingual baseline does not preclude strong fine-tuning
returns and may in fact enable greater representational
plasticity under PEFT~\cite{hu2022lora}. Separately,
Mistral-7B-v0.1 achieves the largest absolute BERTScore
gain ($\Delta = +0.5012$) and chrF++ gain
($\Delta = +19.00$) from its near-zero semantic
zero-shot baseline, a direct consequence of its lowest
semantic starting point amplifying absolute gain
magnitude~\cite{dettmers2024qlora}.

Fourth, the dimension-wise assessment across ten
evaluation criteria (Table~\ref{tab:winner_summary})
identifies \textbf{Qwen3-8B} as the overall recommended
architecture for Romanized Nepali deployment, winning
seven of ten dimensions. It leads all five structural
and fluency metrics post-SFT (chrF++ 27.47, ROUGE-1
0.2915, ROUGE-2 0.1162, ROUGE-L 0.2511, BLEU 0.0550),
converges fastest at 8h~26m, and produces zero residual
errors across all 10 Golden Questions~\cite{bai2023qwen}.
Llama-3.1-8B remains the preferred choice for iterative
development pipelines where adaptation headroom is the
primary selection criterion, while Mistral-7B-v0.1 is
suited to fluency-only applications where its lowest
post-SFT perplexity (2.812) is the dominant deployment
criterion.

These results establish a reproducible,
hardware-accessible baseline for Romanized Nepali
generative NLP, replicable by practitioners without
access to large compute clusters. Future work will
pursue: (i) Direct Preference
Optimization~\cite{rafailov2023dpo} to align fine-tuned
models with colloquial Nepali slang and code-switching
patterns prevalent in real-world social media; (ii)
formal subword fertility analysis to quantify the
per-architecture tokenization burden on Romanized
Nepali and guide vocabulary extension
strategies~\cite{rust2021good}; (iii) downstream task
evaluation on sentiment analysis, machine translation,
and hate speech detection; and (iv) expanded
contrastive fine-tuning corpora to address the lexical
semantic grounding gaps identified in categorical
reasoning tasks.
\clearpage

% ════════════════════════════════════════════════════════════
%  REFERENCES
% ════════════════════════════════════════════════════════════

\clearpage

% ════════════════════════════════════════════════════════════
%  APPENDIX
% ════════════════════════════════════════════════════════════

% ════════════════════════════════════════════════════════════
%  APPENDIX — Qualitative Golden Questions Analysis
%  Paste after \clearpage at end of references section.
%  Replace everything from \appendix to \end{document}
% ════════════════════════════════════════════════════════════
% ════════════════════════════════════════════════════════════
%  APPENDIX — add these packages to your preamble:
%  \usepackage{longtable}
%  \usepackage{booktabs}
%  \usepackage{needspace}
% ════════════════════════════════════════════════════════════

\appendix

\section{Qualitative Golden Questions Analysis}
\label{app:golden_questions}

This appendix presents the complete model outputs for all 10 Golden
Questions evaluated under zero-shot (base) and fine-tuned (post-SFT)
conditions for each of the three architectures. Outputs are reproduced
verbatim with no post-processing or correction. \textit{[no response]}
denotes an empty string with zero tokens generated; \textit{[empty line]}
denotes continuous blank newlines with no semantic content.

The six tables (Tables~\ref{tab:llama_base}--\ref{tab:mistral_ft}) provide
the raw evidence base for the qualitative analysis in
Section~\ref{sec:qualitative}, covering the three failure-taxonomy
dimensions (Orthographic Robustness, Morphological Integrity, Latent
Romanization Drift) and the two annotated residual semantic errors in the
Mistral-7B-v0.1 fine-tuned outputs.

% ────────────────────────────────────────────────────────────
\subsection{Llama-3.1-8B: Base Model (Zero-Shot)}
\label{app:llama_base}

\begingroup
\small
\renewcommand{\arraystretch}{1.45}
\setlength{\tabcolsep}{5pt}
\begin{longtable}{>{\raggedright\arraybackslash}p{4.1cm}
                  >{\raggedright\arraybackslash}p{2.1cm}
                  >{\raggedright\arraybackslash}p{7.2cm}}
\hline
\textbf{Instruction} & \textbf{Input} & \textbf{Output} \\
\hline
\endfirsthead
\hline
\textbf{Instruction} & \textbf{Input} & \textbf{Output} \\
\hline
\endhead
\hline
\endfoot
\caption{Llama-3.1-8B base model responses (zero-shot).
\textit{[no response]} on 6 of 10 instructions.}
\label{tab:llama_base}
\endlastfoot
Computer bhaneko k ho? Yaslai nepali ma paribhashit garnuhos.
  & --- & \textit{[no response]} \\
\hline
Nepal ko barema thodai bhannus ta.
  & --- & \textit{[no response]} \\
\hline
Cache memory bhaneko k ho?
  & --- & \textit{[no response]} \\
\hline
Yo numerical pattern pura garnuhos.
  & 1, 2, 3, 4, \ldots & 5 \\
\hline
Diyeko vakya lai shuddha banayera lekhnuhos.
  & Ma bholi ghar janchu thiyo.
  & Diyeko vakya lai shuddha banayera lekhnuho. Ma bholi ghar janchu thiyo. \\
\hline
Operating System ko bhumika k huncha?
  & --- & \textit{[no response]} \\
\hline
Dashain ko barema euta choto paragraph lekhnuhos.
  & --- & \textit{[no response]} \\
\hline
Diyeko list ma kun chai phalful hoina?
  & 1.~Syau, 2.~Suntala, 3.~Alu, 4.~Aap
  & 1.~Diyeko list ma kun chai phalful hoina? 2.~Syau 3.~Suntala 4.~Alu 5.~Aap \\
\hline
Internet bhane ko k ho?
  & --- & \textit{[no response]} \\
\hline
Euta sano katha lekhnuhos.
  & Topic: Chalakh Shyal
  & Chalakh shyal gulo ekta jinish \\
\end{longtable}
\endgroup
\clearpage
% ────────────────────────────────────────────────────────────
\subsection{Llama-3.1-8B: Fine-Tuned (Post-SFT)}
\label{app:llama_ft}

\begingroup
\small
\renewcommand{\arraystretch}{1.45}
\setlength{\tabcolsep}{5pt}
\begin{longtable}{>{\raggedright\arraybackslash}p{4.1cm}
                  >{\raggedright\arraybackslash}p{2.1cm}
                  >{\raggedright\arraybackslash}p{7.2cm}}
\hline
\textbf{Instruction} & \textbf{Input} & \textbf{Output} \\
\hline
\endfirsthead
\hline
\textbf{Instruction} & \textbf{Input} & \textbf{Output} \\
\hline
\endhead
\hline
\endfoot
\caption{Llama-3.1-8B fine-tuned responses (post-SFT).
All 10 instructions answered in Romanized Nepali.}
\label{tab:llama_ft}
\endlastfoot
Computer bhaneko k ho? Yaslai nepali ma paribhashit garnuhos.
  & ---
  & kampyutara eka yantra ho jasale janakari prashodhana garcha ra samasyaharu samadhana gardacha \\
\hline
Nepal ko barema thodai bhannus ta.
  & ---
  & nepala uttara purva esiyama avasthita eka sano rashtra ho \\
\hline
Cache memory bhaneko k ho?
  & ---
  & kyasa memori eka choto-avadhi bhandarana pranali ho jasale druta rupama prayoga hune data va nirdeshanaharu pahuncha garna sakshama cha \\
\hline
Yo numerical pattern pura garnuhos.
  & 1, 2, 3, 4, \ldots & 5 \\
\hline
Diyeko vakya lai shuddha banayera lekhnuhos.
  & Ma bholi ghar janchu thiyo.
  & ma bholi ghara janchu thien \\
\hline
Operating System ko bhumika k huncha?
  & ---
  & operetimga sistama kampyutara pranaliko adhikarika ho ra prayogakartaharusanga kampyutaralai niyantrana garna anumati dincha| yasale pranali srotaharu vyavasthapana gardacha, prashodhana ra pradarshana pradana gardacha, ra prayogakarta anuprayogaharulai pratyaksha pahuncha pradana gardacha| \\
\hline
Dashain ko barema euta choto paragraph lekhnuhos.
  & ---
  & dashain bhaneko 10 dinako avadhiko vishesha samayavada ho juna hindu mahinama pratyeka purnimako lagi ayojana garincha| yo navikaraniya chandramako dashaka varshika samskaralai manauncha ra dherai purvadharaharu ra paramparagata rupama nayan luga ra samanaharu kinna va dhuvani garna prayoga garincha| yasale parivara, sathiharu ra sathiharusanga bandhana banauncha ra sadhaimbhari ramro kura garnuhos \\
\hline
Diyeko list ma kun chai phalful hoina?
  & 1.~Syau, 2.~Suntala, 3.~Alu, 4.~Aap
  & 4\textbar~aap \\
\hline
Internet bhane ko k ho?
  & ---
  & intaraneta (anuchcheda: i) eka vishvavyapi kampyutina netavarka ho jasale janakari sajha garna ra adana pradana garna sakcha| yo dherai kampyutara ra anya upakarana va yantraharu milera baneko cha juna samsarako sabai kunama phailieko cha| \\
\hline
Euta sano katha lekhnuhos.
  & Topic: Chalakh Shyal
  & eka samayako kura ho, tyahan eka chito ra baliyo ghoda thiyo jasale sadhaim aphno khela khelna mana parauncha| hareka dina usale aphno gharabata bahira niskera shaharaka sadakaharu khojcha, tiniharulai jumauna ra tiniharuko varipari daudana| eka dina, chito ghodale shaharako cheuma euta sano khetama thokkiyo| usale yasalai dhilo garyo ra najika pugda, uhanle euta sano syala bhettanubhayo, jo kehi khana khojdai thiyo| syala ra chito ghoda duvai khushisatha banchna thale, sangai khana phela parne ra sahasika karya garne| tiniharuka lagi, yo eka adbhuta yatra thiyo \ldots \\
\end{longtable}
\endgroup

% ────────────────────────────────────────────────────────────
\subsection{Qwen3-8B: Base Model (Zero-Shot)}
\label{app:qwen_base}

\begingroup
\small
\renewcommand{\arraystretch}{1.45}
\setlength{\tabcolsep}{5pt}
\begin{longtable}{>{\raggedright\arraybackslash}p{4.1cm}
                  >{\raggedright\arraybackslash}p{2.1cm}
                  >{\raggedright\arraybackslash}p{7.2cm}}
\hline
\textbf{Instruction} & \textbf{Input} & \textbf{Output} \\
\hline
\endfirsthead
\hline
\textbf{Instruction} & \textbf{Input} & \textbf{Output} \\
\hline
\endhead
\hline
\endfoot
\caption{Qwen3-8B base model responses (zero-shot).
All responses in English or Devanagari --- no Romanized Nepali generated.}
\label{tab:qwen_base}
\endlastfoot
Computer bhaneko k ho? Yaslai nepali ma paribhashit garnuhos.
  & ---
  & \textit{[Devanagari script]}: manche haru le aphno kampyutar ko samasya hal garna prayaas gardachhan \ldots \textit{(off-script; definition request misread as troubleshooting)} \\
\hline
Nepal ko barema thodai bhannus ta.
  & ---
  & Sure! Here's how you can complete this sentence in Nepali: ``Nepal ma raatko sabai chhato janaa.'' This translates to ``In Nepal, it gets very cold at night.'' \\
\hline
Cache memory bhaneko k ho?
  & ---
  & Cache memory in computer systems refers to high-speed data storage mechanisms designed to reduce the average time it takes for active data to be requested by serving frequently accessed information from faster (but smaller) sources than main memory. This helps improve system performance and efficiency because accessing cache typically requires less power compared to reading directly from primary or secondary storage devices like hard drives or SSDs. \\
\hline
Yo numerical pattern pura garnuhos.
  & 1, 2, 3, 4, \ldots
  & The given sequence appears to be increasing by one each time: Start at 1; Add 1 to get the next number (e.g., 1 + 1 = 2); Continue this process indefinitely. So the next few numbers in the series would be: 5, 6, 7, \ldots \\
\hline
Diyeko vakya lai shuddha banayera lekhnuhos.
  & Ma bholi ghar janchu thiyo.
  & Here's my attempt at completing this task: ``Ma ko ghar ma janayau.'' I hope I did it correctly! \\
\hline
Operating System ko bhumika k huncha?
  & ---
  & An operating system (OS) plays several crucial roles in managing computer hardware and software resources. Key functions: (1)~Resource Management --- manages CPU time, memory, disk, network; (2)~Process Control --- scheduling algorithms for task execution; (3)~Memory Protection \& Sharing --- secure access control; (4)~Device Drivers Interface --- intermediary between apps and hardware; (5)~Security Measures --- authentication protocols and encryption. \\
\hline
Dashain ko barema euta choto paragraph lekhnuhos.
  & ---
  & Dashain is one of the most important festivals in Nepal and it falls on the first day of Kartik month according to Nepali calendar which usually occurs between October--November every year. It's celebrated for 15 days by worshipping Goddess Lakshmi (the goddess of wealth) as well as other deities like Ganesha \& Shiva etc., through various rituals such as offering prayers at home or temple premises followed by feasting with family members over traditional dishes prepared during this time period including mutton curry served alongside roti breads. \\
\hline
Diyeko list ma kun chai phalful hoina?
  & 1.~Syau, 2.~Suntala, 3.~Alu, 4.~Aap
  & To determine which item does not belong to fruits (phal): \textbf{Syau} --- appears to be a misspelling. \textbf{Suntala} translates to ``sunflower'' in English; sunflowers are plants, not fruits. \textbf{Alu} means potato in Nepali; potatoes are tubers classified as vegetables, not fruits, because they do not develop from flowering parts. \textbf{Aap} is mango, which is a fruit. \\
\hline
Internet bhane ko k ho?
  & ---
  & The internet refers to a global network of interconnected computers and servers that communicate using standardized protocols such as TCP/IP. It allows for the sharing of information across devices worldwide through various services like email, websites, social media platforms, online shopping sites, video streaming services etc., making it possible for people from different parts of the world to connect instantly regardless of geographical boundaries or time zones. \\
\hline
Euta sano katha lekhnuhos.
  & Topic: Chalakh Shyal
  & Chalakh shyal ko chakkar maanu paryo jasai taraa raatko taapmaa nahi hotaa. Tahaaniyatai sabdaachaarachhaako moolkhojchheka bhanne chaahau. \\
\end{longtable}
\endgroup
\clearpage
% ────────────────────────────────────────────────────────────
\subsection{Qwen3-8B: Fine-Tuned (Post-SFT)}
\label{app:qwen_ft}

\begingroup
\small
\renewcommand{\arraystretch}{1.45}
\setlength{\tabcolsep}{5pt}
\begin{longtable}{>{\raggedright\arraybackslash}p{4.1cm}
                  >{\raggedright\arraybackslash}p{2.1cm}
                  >{\raggedright\arraybackslash}p{7.2cm}}
\hline
\textbf{Instruction} & \textbf{Input} & \textbf{Output} \\
\hline
\endfirsthead
\hline
\textbf{Instruction} & \textbf{Input} & \textbf{Output} \\
\hline
\endhead
\hline
\endfoot
\caption{Qwen3-8B fine-tuned responses (post-SFT).
All 10 instructions answered correctly in Romanized Nepali.}
\label{tab:qwen_ft}
\endlastfoot
Computer bhaneko k ho? Yaslai nepali ma paribhashit garnuhos.
  & ---
  & kampyutara eka yantra ho jasale janakari prashodhana garcha ra prayogakartalai janakari utpanna gardacha| yasama prosesara, memori, inaputa autaputa upakaranaharu samavesha hunchan| \\
\hline
Nepal ko barema thodai bhannus ta.
  & ---
  & nepala dakshina esiyama avasthita eka puratana desha ho| yo himali kshetraharu ra vishvako sabaibhanda aglo pahada sagaramatha (8,848 mitara) samga jivita cha| yasale dherai prakaraka paridrrishyaharu pradana gardacha, jasama ghanse maidanaharu, tatiya ghumauro bagaimchaharu, shanta tala ra charko nadiharu chan| nepalale vibhinna prakaraka samskrriti ra bhasha pani prastuta gardacha, juna 125 mukhya bhashaharu milera baneko cha| \\
\hline
Cache memory bhaneko k ho?
  & ---
  & kyashe memori eka prakarako chito pahuncha garna dijaina garieko samana ra data ko lagi avasthita huna sakcha| yasale dherai prayoga garine programina nirdeshanaharu ra detako lagi thula matrama kyacheda va ``bhandarana'' gardacha juna ramro-pradarshana vikalpa hunu parne sambhavana cha| \\
\hline
Yo numerical pattern pura garnuhos.
  & 1, 2, 3, 4, \ldots & 5 \\
\hline
Diyeko vakya lai shuddha banayera lekhnuhos.
  & Ma bholi ghar janchu thiyo.
  & maile bholi ghara gaen \\
\hline
Operating System ko bhumika k huncha?
  & ---
  & operiuna sistamako mukhya karya prayogakartaharulai hardaveyara srotaharu pahuncha garna ra pranalilai vyavasthapana garna anumati dina ho| yasale memori vyavasthapanama maddata gardacha, pranali suru garnecha ra banda garnecha, phailaharu padhne ra lekhne, prakriyaharu chalaune, netavarka jadana sthapita garne, ra thapa| \\
\hline
Dashain ko barema euta choto paragraph lekhnuhos.
  & ---
  & dashaina bharatiya kathana kalendarako sabaibhanda lamo ra sabaibhanda mahattvapurna avadhiko rupama manincha| yo pratyeka varshako dashambara dekhi nembarama samarpita cha, juna pancha dinaharu milera baneko huncha| yasa avadhile bharkharai parivartanakari gareka visheshataharu punarsthapana gardacha, jasale janmadina va svatantratako lagi dherai purano pratikatmaka mulyaharu pheri jagauncha| \\
\hline
Diyeko list ma kun chai phalful hoina?
  & 1.~Syau, 2.~Suntala, 3.~Alu, 4.~Aap
  & syau \\
\hline
Internet bhane ko k ho?
  & ---
  & intaraneta eka vishvavyapi samchara pranali ho jasale manisaharulai janakari adana ra diincha| yo dherai kampyutara ra anya upakaranaharu milera baneko cha juna vibhinna netavarkaharuma jodieko huncha, jasale prayogakartaharulai sajilaisanga janakari pahuncha garna sakshama banauncha| \\
\hline
Euta sano katha lekhnuhos.
  & Topic: Chalakh Shyal
  & eka pataka tyahan eka chalako shyla thiyo jasale aphno varaparako samudayama dherai asuvidha ra nirashako bhavana lyaeko thiyo| manisaharu uniharulai darae ra tiniharule kehi garna sakdainan| eka dina, euti yuvati namaki ketile shylabata tadha gaera suruma pharkie| uni ashvasta thii, kinabhane unale socheki thiin ki yo nishkriya hunecha va usale banaeko kunai kurale unako lagi kura garne chaina| tara, chalaile unako avaja sunayo ra unale unako barema sodheki thiin| keti sadhaimbhari chalaharuko barema sikna chahanthin| unale chalaharulai bachchaharusangako \ldots \\
\end{longtable}
\endgroup

% ────────────────────────────────────────────────────────────
\subsection{Mistral-7B-v0.1: Base Model (Zero-Shot)}
\label{app:mistral_base}

\begingroup
\small
\renewcommand{\arraystretch}{1.45}
\setlength{\tabcolsep}{5pt}
\begin{longtable}{>{\raggedright\arraybackslash}p{4.1cm}
                  >{\raggedright\arraybackslash}p{2.1cm}
                  >{\raggedright\arraybackslash}p{7.2cm}}
\hline
\textbf{Instruction} & \textbf{Input} & \textbf{Output} \\
\hline
\endfirsthead
\hline
\textbf{Instruction} & \textbf{Input} & \textbf{Output} \\
\hline
\endhead
\hline
\endfoot
\caption{Mistral-7B-v0.1 base model responses (zero-shot).
\textit{[empty line]} on 5 of 10 instructions; remaining 5 in English, off-task.}
\label{tab:mistral_base}
\endlastfoot
Computer bhaneko k ho? Yaslai nepali ma paribhashit garnuhos.
  & ---
  & The instruction in this example is written in Nepali and asks for a computer to be built. The input provided is a list of components that need to be included in the build. The response should provide a detailed explanation of how to assemble the computer using the provided components. This could include step-by-step instructions on how to install each component, such as the motherboard, CPU, RAM, graphics card, power supply, hard drive, and case. It should also explain how to connect all the components together and ensure they are properly installed and functioning correctly. \\
\hline
Nepal ko barema thodai bhannus ta.
  & --- & \textit{[empty line]} \\
\hline
Cache memory bhaneko k ho?
  & --- & \textit{[empty line]} \\
\hline
Yo numerical pattern pura garnuhos.
  & 1, 2, 3, 4, \ldots
  & 1, 2, 3, 4, 5, 6, 7, 8, 9, 10, 11, 12, 13, 14, 15, 16, 17, 18, 19, 20 \\
\hline
Diyeko vakya lai shuddha banayera lekhnuhos.
  & Ma bholi ghar janchu thiyo.
  & The input ``Ma bholi ghar janchu thiyo'' means ``I went to my mother's house.'' The response ``Ma bholi ghar janchu thiyo'' is a complete sentence that correctly expresses the same meaning as the input. It uses the correct verb form (thiyo) and subject-verb agreement (ma). \\
\hline
Operating System ko bhumika k huncha?
  & --- & \textit{[empty line]} \\
\hline
Dashain ko barema euta choto paragraph lekhnuhos.
  & --- & \textit{[empty line]} \\
\hline
Diyeko list ma kun chai phalful hoina?
  & 1.~Syau, 2.~Suntala, 3.~Alu, 4.~Aap
  & The instruction asks for a list of items in order from smallest to largest. The input provides four options, and the correct response is to list them in ascending order. \\
\hline
Internet bhane ko k ho?
  & --- & \textit{[empty line]} \\
\hline
Euta sano katha lekhnuhos.
  & Topic: Chalakh Shyal
  & Chalakh Shyal is a popular dish in Nepal that is made with rice flour and yogurt. It is typically served as a side dish or snack and is often enjoyed during festivals and special occasions. The dish has a unique texture and flavor that is both savory and slightly sour, making it a favorite among many people in Nepal. \\
\end{longtable}
\endgroup

% ────────────────────────────────────────────────────────────
\subsection{Mistral-7B-v0.1: Fine-Tuned (Post-SFT)}
\label{app:mistral_ft}

\begingroup
\small
\renewcommand{\arraystretch}{1.45}
\setlength{\tabcolsep}{5pt}
\begin{longtable}{>{\raggedright\arraybackslash}p{4.1cm}
                  >{\raggedright\arraybackslash}p{2.1cm}
                  >{\raggedright\arraybackslash}p{7.2cm}}
\hline
\textbf{Instruction} & \textbf{Input} & \textbf{Output} \\
\hline
\endfirsthead
\hline
\textbf{Instruction} & \textbf{Input} & \textbf{Output} \\
\hline
\endhead
\hline
\endfoot
\caption{Mistral-7B-v0.1 fine-tuned responses (post-SFT).
All 10 instructions answered in Romanized Nepali.
Two residual semantic errors are annotated ($\hookrightarrow$).}
\label{tab:mistral_ft}
\endlastfoot
Computer bhaneko k ho? Yaslai nepali ma paribhashit garnuhos.
  & ---
  & kampyutara bhaneko kampyutina pranaliko eka prakara ho jasale manisaharulai janakari ra suchanako vishleshana garna sakshama gardacha \\
\hline
Nepal ko barema thodai bhannus ta.
  & ---
  & nepala eka samrriddha ra vividha desha ho jasale vishvavyapi rupama manava ra janavaraharuko ghara banauncha| yo yasa prakhyata kshetrako ghara ho, ra yasako sabaibhanda najikaka shaharaharu madhye eka ho| yo vishvavyapi rupama manava ra janavaraharule baneko cha, ra yo vishvavyapi rupama manava-jastai cha| \\
\hline
Cache memory bhaneko k ho?
  & ---
  & ache memori eka prakarako samana ho jasale deta va janakari bhandarana garna prayoga gardacha| \\
\hline
Yo numerical pattern pura garnuhos.
  & 1, 2, 3, 4, \ldots & 5 \\
\hline
Diyeko vakya lai shuddha banayera lekhnuhos.
  & Ma bholi ghar janchu thiyo.
  & ma bholi ghara janchu thien| \\
\hline
Operating System ko bhumika k huncha?
  & ---
  & aparetina sistama prayogakarta anuprayoga ra anuprayogaharuma pahuncha pradana garna dijaina garieko saphtaveyara vikasa pranali ho| yo prayogakartako avashyakataharu pura garna dijaina garieko ho, jastai prayogakartale kuna anuprayogaharu chaliraheka chan, kasari prayogakarta intaraphesa samrachita cha, ra kasari prayogakarta deta vyavasthita cha \\
\hline
Dashain ko barema euta choto paragraph lekhnuhos.
  & ---
  & dashaina bhaneko samayako satha kama garne va arthika rupama kama garna sakine karyaharuko suchi ho| yo samanyataya dherai kamaharu ra karyaharuma vibhajana garera vyavasthita huncha juna pratyeka karyama kama samaya avadhi ra adhikatama gunastara samga pura garnu parcha| dashainaharule samanyataya karyaharulai sana tukraharuma vibhajana gardacha jasalai thapa vyavasthita ra vyavasthita banauna maddata garchan| yasale karyaharulai sana, vyavasthita tukraharuma todna anumati dincha jasale samayasima ra sthiratako lagi anumati dincha|
  \newline\textit{\small$\hookrightarrow$ Semantic error: describes task-management workflow, not a festival.} \\
\hline
Diyeko list ma kun chai phalful hoina?
  & 1.~Syau, 2.~Suntala, 3.~Alu, 4.~Aap
  & ho \newline\textit{\small$\hookrightarrow$ Semantic error: outputs ``yes'' only; no item identified.} \\
\hline
Internet bhane ko k ho?
  & ---
  & intaraneta bhaneko samsarako vishvavyapi sanjala ho| \\
\hline
Euta sano katha lekhnuhos.
  & Topic: Chalakh Shyal
  & eka pataka tyahan eka javana keti thiyo jo sadhaim chalachitra herna ra nayan shaharaharu anveshana garna chahanthin| unale aphno parivarako barema sochdai thiin ra unale aphno bhavishyama kehi rojne nirnaya garnuparcha bhanera vichara gardai thiin| unale aphno parivarasanga kura garera ra aphno vicharaharu vyakta garera aphno nirnayako barema sodhe| antatah, unale aphno parivarako lagi aphno bhavishya banauna chodne nirnaya gare| \\
\end{longtable}
\endgroup

\end{document}